\documentclass[autheryear, times, 9pt]{elsarticle}

\usepackage{amsmath}
\usepackage{amsthm}
\usepackage{amssymb}
\usepackage[linesnumbered,ruled]{algorithm2e}
\usepackage{color}
\usepackage{diagbox}
\usepackage{float}
\usepackage{geometry}
\usepackage{graphicx}
\usepackage{hyperref}
\usepackage{longtable}
\usepackage{multirow}
\usepackage{makecell}
\usepackage{slashbox}
\usepackage{soul}
\usepackage{subfigure}

\usepackage{url}
\usepackage{verbatim}
\usepackage{hyperref}

\restylefloat{figure}

\newtheorem{assumption}{Assumption}

\newcommand{\Mu}{\mathcal{U}}
\newcommand{\Nu}{\mathcal{V}}
\newcommand{\Rho}{\mathcal{P}}
\newcommand{\Alpha}{\mathcal{A}}
\usepackage{natbib}

\hypersetup{
    colorlinks,
    linkcolor={red!50!black},
    citecolor={blue!50!black},
    urlcolor={blue!80!black}
}

\biboptions{authoryear}

\DeclareMathOperator*{\argmax}{argmax}

\setcitestyle{notesep={; },round,aysep={},yysep={;}}
\journal{Transportation Research Part C: Emerging Technologies}

\begin{document}
\begin{frontmatter}
\title{Demonstration-guided Deep Reinforcement Learning for Coordinated Ramp Metering and Perimeter Control in Large Scale Networks}
\author[address1]{Zijian Hu}
\ead{zijian.hu@connect.polyu.hk}
\author[address1,address4]{Wei Ma\corref{mycorrespondingauthor}}
\cortext[mycorrespondingauthor]{Corresponding author}
\ead{wei.w.ma@polyu.edu.hk}
\address[address1]{Civil and Environmental Engineering, The Hong Kong Polytechnic University \\ Hung Hom, Kowloon, Hong Kong SAR, China}
\address[address4]{Research Institute for Sustainable Urban Development, The Hong Kong Polytechnic University \\ Hung Hom, Kowloon, Hong Kong SAR, China}

\begin{abstract}
Effective traffic control methods have great potential in alleviating network congestion. Existing literature generally focuses on a single control approach, while few studies have explored the effectiveness of integrated and coordinated control approaches.
This study considers two representative control approaches: ramp metering for freeways and perimeter control for homogeneous urban roads, and we aim to develop a deep reinforcement learning (DRL)-based coordinated control framework for large-scale networks. The main challenges are 1) there is a lack of efficient dynamic models for both freeways and urban roads; 2) the standard DRL method becomes ineffective due to the complex and non-stationary network dynamics.
In view of this, we propose a novel meso-macro dynamic network model and first time develop a demonstration-guided DRL method to achieve large-scale coordinated ramp metering and perimeter control. 
The dynamic network model hybridizes the link and generalized bathtub models to depict the traffic dynamics of freeways and urban roads, respectively. 
For the DRL method, we incorporate demonstration to guide the DRL method for better convergence by introducing the concept of ``teacher'' and ``student'' models. The teacher models are traditional controllers (\textit{e.g.}, ALINEA, Gating), which provide control demonstrations. 
The student models are DRL methods, which learn from the teacher and aim to surpass the teacher’s performance. 
To validate the proposed framework, we conduct two case studies in a small-scale network and a real-world large-scale traffic network in Hong Kong. Numerical results show that the proposed DRL method outperforms demonstrators as well as the state-of-the-art DRL methods, and the coordinated control is more effective than just controlling ramps or perimeters by 8.7\% and 22.6\%, respectively. The research outcome reveals the great potential of combining traditional controllers with DRL for coordinated control in large-scale networks.
\end{abstract}

  
\end{frontmatter}

\section{Introduction}
With rapid urbanization and city agglomeration, traffic congestion has become one of the most significant urban challenges in recent years. Intelligent and effective control strategies are in great need to alleviate congestion issues for public agencies.
Particularly, we notice that the structure of urban road networks is in nature heterogeneous, and it consists of different types of roads ({\em e.g.}, freeways and local roads) that function differently for mobility and accessibility purposes. Various control and management strategies should be enacted adaptively according to different types of roads. For instance, ramp metering is a common control method for freeways by controlling the vehicles merged into freeways from ramps \citep{alinea,pialinea}, and it maintains the smoothness of the mainstream traffic by preventing the capacity drop caused by upstream queues \citep{capacity_drop01,capacity_drop02}. In contrast, perimeter control focuses on urban roads by restricting the inflow of vehicles into congested areas \citep{pi1_pm,pi2_pm}, where the traffic dynamics can be modeled using the Macroscopic Fundamental Diagram (MFD) \citep{MFD01,MFD02}. However, how to control the traffic for heterogeneous road types is still an open question \citep{Haddad_PC}.

The scale of networks also limits the performance of current control methods. Previously, conventional controllers (\textit{e.g.}, Proportional–Integral–Derivative (PID) \citep{alinea, pialinea, pi1_pm}, Model Predictive Control (MPC) \citep{mpc_rm, mpc1_pm}, \textit{etc.}) are leveraged to obtain the optimal policy on local road networks. However, these methods may require delicate calibrations in large-scale networks due to the exponentially growing problem scale and complexity. Additionally, the heterogeneity of road types on a large-scale network makes it complicated to model the network traffic dynamics \citep{hu2022towards}, which also impedes the effectiveness of existing traffic control methods \citep{qian2016dynamic}. 

In recent years, Deep Reinforcement Learning (DRL) draws attention in the research communities as it has been validated as an effective method to solve large-scale and complex control problems \citep{su2023hierarchical, zhou2023scalable}. In DRL methods, agents learn knowledge by exploring unknown environments. An agent is usually modeled with a finetuned neural network that outputs an action based on its own policy and observation. The DRL methods also demonstrate great potential in ramp metering \citep{rm_q_learning_co,rm_equity,marl_rm} and perimeter control \citep{zhou1_pm,zhou2_pm}, respectively.
However, when the size of networks becomes large and the control scenario becomes complex, the non-stationary environment and large search spaces may also prevent the DRL from finding the optimal control policy. The non-stationary environment refers to the situation where multiple agents ({\em i.e.}, ramps and perimeters) may have various control policies, and the large searching spaces are due to the size of networks.

\begin{figure}[h]
    \centering
    \includegraphics[width=0.8\textwidth]{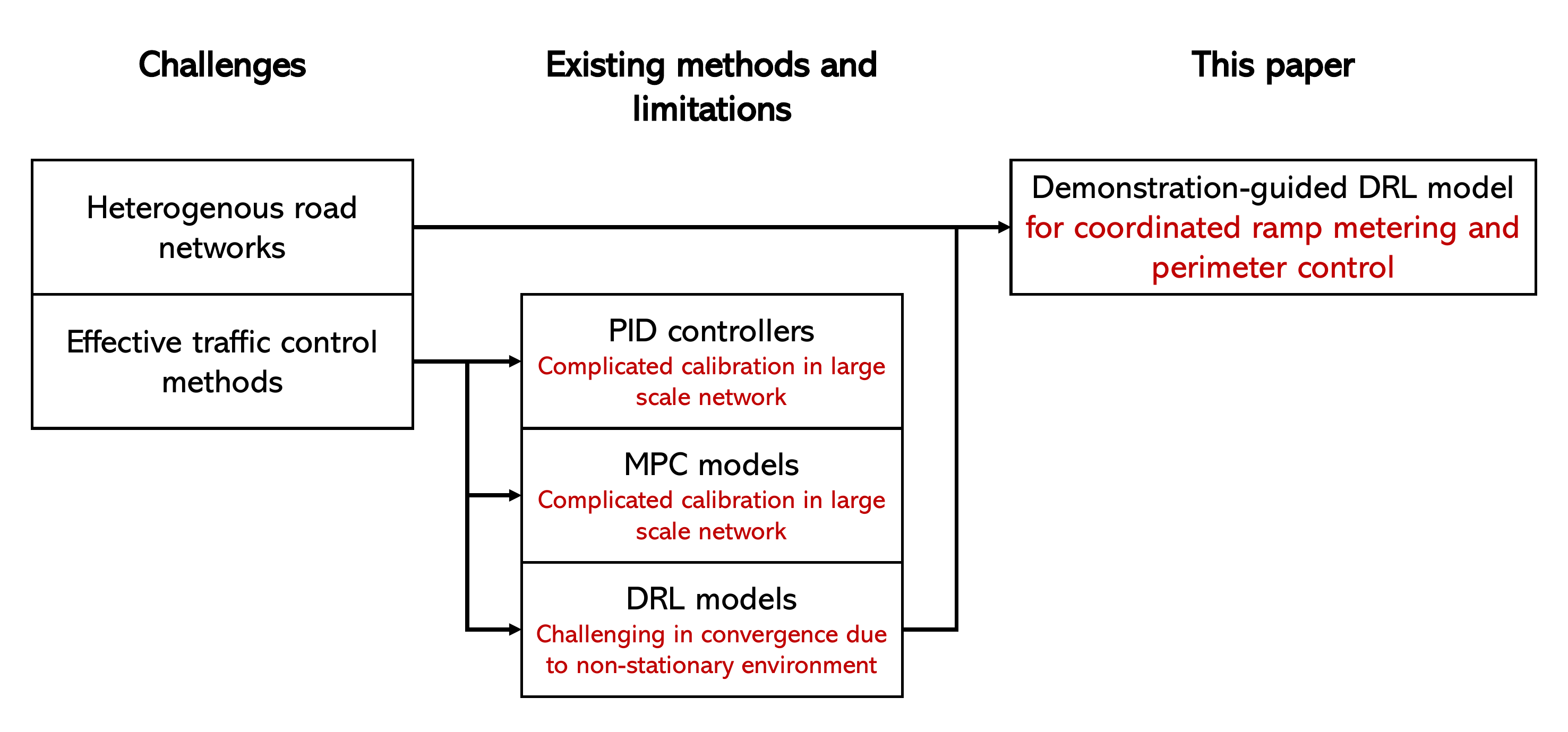}
    \caption{Challenges of existing methods for network-wide coordinated traffic control.}
    \label{fig:challenges}
\end{figure}

The aforementioned challenges in the network-wide coordinated traffic control are summarized in Figure~\ref{fig:challenges}.

In this paper, we propose a novel idea to allow the DRL not only to learn by explorations from the environment but also to learn from a teacher. Conceptually, a teacher ({\em i.e.}, demonstrator) can provide examples to show how to accomplish a task, and the student ({\em i.e.}, our DRL agent) can use these demonstrated examples to learn knowledge through imitation and generalization. Similar ideas have been applied to games through DRL. For example, the AlphaGo algorithm \citep{alphago}, which learns from a combination of human expert games and self-play, can defeat the human Go champion. In the StarCraft game, the AlphaStar algorithm can reach the top human players by learning the game records \citep{alphastar}. However, the idea of using demonstration guidance in engineering applications, especially in traffic control problems, has been rarely explored \citep{demon_gen_01,demon_gen_02,demon_signal_01}. Hence this paper aims to fill the gap and develop the demonstration-guided DRL for both ramp metering and perimeter control.

For most existing demonstration-guided DRL methods, the demonstrators are human. For example, the autonomous vehicles can learn the driving behaviors from human drivers \citep{demon_av_01,demon_av_02}. However, it is difficult to enact a decent control policy even for a human in perimeter control and ramp metering. To this end, we adopt the traditional decentralized controllers to provide sub-optimal policies for large-scale network control, and later we will show that the student can actually outperform the teachers by integrating demonstrations and explorations.

Additionally, there is a lack of existing models that depict the traffic flow dynamics on large-scale networks with heterogeneous road types. To this end, we develop a meso-macro traffic model that integrates the mesoscopic and macroscopic dynamic traffic models for freeways and local roads, respectively. The mesoscopic link model (\textit{e.g.}, Asymmetric Cell Transmission Model (ACTM) \citep{actm}) depicts the flow dynamics at freeways, on-ramps, and off-ramps; and the macroscopic bathtub model manipulates vehicles in  homogeneous regions \citep{bathtub}. Both models have been hybridized to accommodate different control scenarios in our previous studies \citep{hu2022towards}, while further studies are required for DRL-based traffic control.




To summarize, this paper focuses on the coordinated ramp metering and perimeter control on large-scale networks. To model the traffic flow dynamics on heterogeneous networks, we propose a meso-macro traffic model using ACTM and generalized bathtub models. On top of that, we develop a demonstration-guided DRL method that is trained by incorporating ``teacher'' and ``student'' models. The teacher models are traditional controllers (\textit{e.g.}, ALINEA), which could provide imperfect control demonstrations. The student models are DRL methods, which learn from the teacher and aim to surpass the teacher's performance. To validate the proposed framework, we conducted two case studies in a small-scale network and a real-world large-scale network in Hong Kong. 
The proposed method outperforms the baseline methods of existing traditional and DRL controllers in both case studies. 

The major contributions of this paper are as follows: 
\begin{itemize}
    \item It formulates the coordinated ramp metering and perimeter control in a large-scale network as a DRL problem.
    \item It develops a novel meso-macro traffic model that supports large-scale dynamic network loading. The proposed model integrates ACTM and the generalized bathtub model for freeways and local road networks, respectively.
    \item It first time proposes the concept of demonstration for DRL-based traffic control, and it develops a DRL method guided by demonstrators to improve the performance of coordinated ramp metering and perimeter control.
    \item It conducts coordinated control experiments on both a small-scale and a real-world large-scale network in Hong Kong. The proposed method outperforms the traditional controllers and state-of-the-art DRL methods.
\end{itemize}

The rest of this paper is organized as follows. Section \ref{sec:literature} discusses the literature related to this study. Section \ref{sec:methods} presents the proposed framework of the dynamic traffic modeling and the DRL method. In section \ref{sec:experiments}, numerical experiments at a small-scale and a real-world large-scale network in Hong Kong are presented. Finally, conclusions are drawn in section \ref{sec:conclusion}. All notations used in this paper are summarized in \ref{apx:notations}.

\section{Literature review}
\label{sec:literature}
This section reviews the literature on related topics in this paper.

\subsection{Dynamic traffic models}
Based on the modeling scale and resolution, the dynamic traffic models can be categorized into 
the microscopic, mesoscopic, and macroscopic models \citep{traffic_models}. 
Microscopic models depict vehicle acceleration and deceleration based on the interactions among vehicles nearby \citep{OpenAI_Gyms,Flow,SUMO,CityFlow,BARK,SMARTS,Aimsun,CBEngine,Vissim}. However, due to the model complexity and traffic modeling resolution, it is nearly impossible to operate in a large-scale network. Most mesoscopic models the traffic dynamics based on vehicle queues, which simplifies the modeling of individual vehicles and increases the modeling efficiency \citep{DynaMIT,DTALite,POLARIS,MATSim,SUMO,Vissim}. However, it is still challenging to achieve a large-scale traffic modeling due to the gridlock effect \citep{pm_first}, which is a special case of the traffic congestion in which vehicles are blocked in a circular queue. 
The gridlock is often triggered by the improper setting of link attributes in queue-based models. 
The aforementioned traffic models are link-based models, and recently, the macroscopic network-based traffic models draw attention in the research community \citep{MFD01,MFD02}. The generalized bathtub model, a macroscopic network-based model, allows large-scale modeling, as well as decent traffic information \citep{bathtub}. In this paper, the macroscopic network-based model is referred to as the macroscopic model, while we omit the link-based macroscopic model as it is not suitable for dynamic traffic control.
Generally, it would be beneficial to integrate the advantages of mesoscopic and macroscopic network models for large-scale networks. 

\subsection{Reinforcement Learning (RL) and Deep Reinforcement Learning (DRL)}
RL has been extensively studied for solving Markovian decision-making problems \citep{sutton}. Traditional RL methods solve the control problem by updating a value table until convergence \citep{td, sarsa, qlearning}. However, the complexity of such methods grows exponentially with respect to the problem complexity. With the development of Deep Learning (DL), the combination of RL and DL is a promising direction for solving high-dimensional control problems. For the discrete control, Deep Q Network (DQN) is the first DRL framework playing Atari games, which outperformed human experts. Several variants have been proposed to refine the framework \citep{double_dqn, duel_dqn, prb_dqn, dis_dqn, noisy_dqn, rainbow_dqn}. For continuous control, Deterministic Deep Policy Gradient (DDPG) extends the DQN model with the actor-critic framework to solve the continuous control problem \citep{ddpg}. Asynchronous Advantage Actor Critic (A3C) model could accelerate the training process by incorporating distributed sampling strategy \citep{a3c}. Proximal Policy Optimization (PPO) \citep{ppo} and Soft Actor-Critic (SAC) \citep{sac} are advanced DRL methods for continuous control. However, in a multi-agent system, the non-stationary environment may disturb the convergence of DRL methods, resulting in poor control performance. In this study, we incorporate the demonstrated-guided DRL method to enhance the performance of coordinated ramp metering and perimeter control.

\subsection{Ramp metering and perimeter control}
Ramp metering serves as an important and effective freeway control strategy that has been studied in the past few decades. Previously, ramp metering can be achieved through Proportional Integral (PI) controllers \citep{alinea,pialinea,metaline,hero} and MPC models \citep{mpc_rm}. Nowadays, DRL methods gradually draw attention as they succeeded in many complex control problems. For example, \citet{rm_q_learning} proposed an RL-based local ramp control strategy using the Q-Learning algorithm. In the follow-up work, they generalized the Q-Learning algorithm into a coordinated control with multiple ramps by integrating ramps as a centralized agent \citep{rm_q_learning_co}. Centralized control of ramps can work when the ramp number is small. However, it is challenging to simultaneously control tens of ramps on the entire freeway since the space of joint action distribution will grow exponentially. To solve this problem, \citet{marl_rm} used mutual weight regularization to reduce the impact of the high dimension of action space in multi-ramp control. \citet{mappo_rm} leveraged Multi-Agent Proximal Policy Optimization (MAPPO) algorithm to improve the control performance. \citet{rm_equity} considered user equity when controlling ramps on freeways. A recent work incorporated physical-informed RL-based ramp metering using a combination of historical data and synthetic data generated from traffic models \citep{offline_rl_rm}.

Perimeter control, first proposed by \citet{pm_first}, is an effective control method for urban regions. \citet{lqr_pm} introduced a multi-region perimeter control method using Linear Quadratic Regulator (LQR), while several studies incorporated the Proportional Integral (PI) controllers for perimeter control of urban areas \citep{pi1_pm, pi2_pm, pi3_pm, pi4_pm}. There are other methods such as MPC-based methods \citep{mpc1_pm, mpc2_pm, mpc3_pm, mpc4_pm}, dynamic programming \citep{zichengsu}, and model-based RL methods \citep{canchen} for multi-region perimeter control. Recent studies focused on data-driven and model-free adaptive control since they do not need accurate modeling of the traffic flow system, which may improve the performance in the real-world application \citep{mfac2_pm,mfac1_pm, mfac3_pm,zhou2023scalable}. For DRL methods, \citet{rl_pm} proposed an RL-based for the perimeter control of a single region using the DQN model. \citet{zhou1_pm} considered the discrete and continuous control of two regions, respectively, using Double DQN and  DDPG model. They further incorporated domain control knowledge into the DRL framework to accommodate complex urban network \citep{zhou2_pm}.

To the best of our knowledge, the topic of coordinated ramp metering and perimeter control has not been fully explored. \citet{Haddad_PC} initially considered the cooperative control of two regions and a freeway using perimeter control and ramp metering. \citet{CRCPC} proposed an integral control method for the joint control of ramps and perimeters in a real-world road network. \citet{VSLPC} introduced a joint control method to combine the Variable Speed Limit (VSL) and perimeter control for mixed traffic networks. In this paper, we consider the large-scale road network and focus on the joint effects of ramp metering and perimeter control.

\section{Model}
\label{sec:methods}
This section discusses the proposed framework in detail. We first present a meso-macro dynamic network model to depict the network-wide traffic dynamics, followed by the development of the demonstration-guided DRL method.

\begin{figure}[h]
    \centering
    \includegraphics[width=0.9\textwidth]{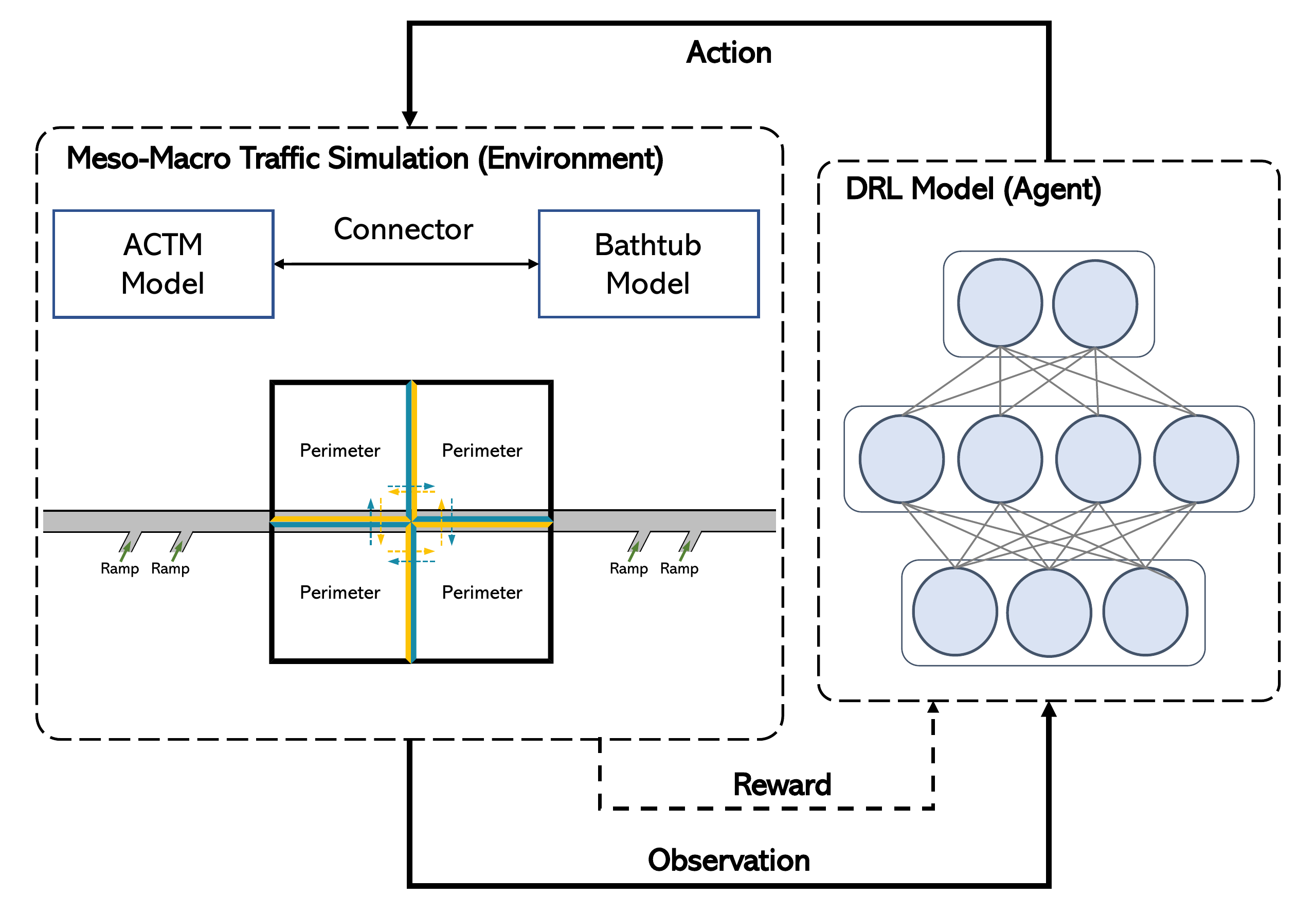}
    \caption{The framework of the coordinated ramp metering and perimeter control.}
    \label{fig:framework}
\end{figure}

The overall framework of the coordinated ramp metering and perimeter control is shown in Figure~\ref{fig:framework}. For the control problem using DRL, there are two essential modules: the environment and the agent. 
An environment is an object that a decision-maker will interact with to achieve a certain goal. 
The decision maker is viewed as an agent. The agent can observe the environment, and it can respond to the environment by taking an action based on its observations. Additionally, the agent will receive reward feedback to assess how well the action is. The agent will keep interacting with the environment to maximize its total reward. DRL concentrates on finding the optimal control policy for the agent in finite steps. For the specialized topic in this paper, an agent is either a ramp or a perimeter, the environment is the meso-macro dynamic traffic model, and the goal is to minimize Total Travel Time (TTT) for all travelers ({\em i.e}, System Optimal).



\subsection{Meso-macro dynamic traffic model}
The environment is depicted by the meso-macro dynamic traffic model for large-scale networks. 
The proposed dynamic traffic model contains two fundamental elements, the ACTM and the generalized bathtub model. The ACTM models the traffic dynamics on freeways, on-ramps, and off-ramps, and vehicles are ruled by the bathtub model in urban areas. Both models are adapted for the DRL-based traffic control. 
A preliminary version of this model is developed in our previous work \citep{hu2022towards}, and we further extend it to model the effects of ramp metering and perimeter control.

The road network is formulated as a directed graph $\mathbf{G} = \langle \mathbf{V}, \mathbf{E} \rangle$, where $\mathbf{V}$ represents the node set (\textit{i.e.,} merges, diverges and general junctions), and $\mathbf{E}$ represents the link set (\textit{i.e.,} road with different types and lanes). According to the netwrok attributes, the whole network $\mathbf{G}$ can be divided into various subgraphs $\mathbf{G}_1, \mathbf{G}_2, \dots, \mathbf{G}_d, \dots, \mathbf{G}_{\mathcal{N}_d} \subseteq \mathbf{G}$, where $\mathbf{G}_d = \langle \mathbf{V}_d, \mathbf{E}_d\rangle$, $\{ \mathbf{V}_1, \mathbf{V}_2, \dots, \mathbf{V}_d, \dots \mathbf{V}_{\mathcal{N}_d} \}$ is a partition of set $\mathbf{V}$. $\mathcal{N}_d$ means the number of partitioned regions. Each subgraph represents a region. Without loss of generality, we assume regions $\mathbf{G}_1, \mathbf{G}_2, \dots, \mathbf{G}_{\mathcal{N}_d - 1}$ represent urban areas and are modeled by bathtub models; region $\mathbf{G}_{\mathcal{N}_d}$ presents the freeway network and is modeled by ACTM.

\subsubsection{ACTM}
\begin{figure}[h]
    \centering
    \includegraphics[width=1\textwidth]{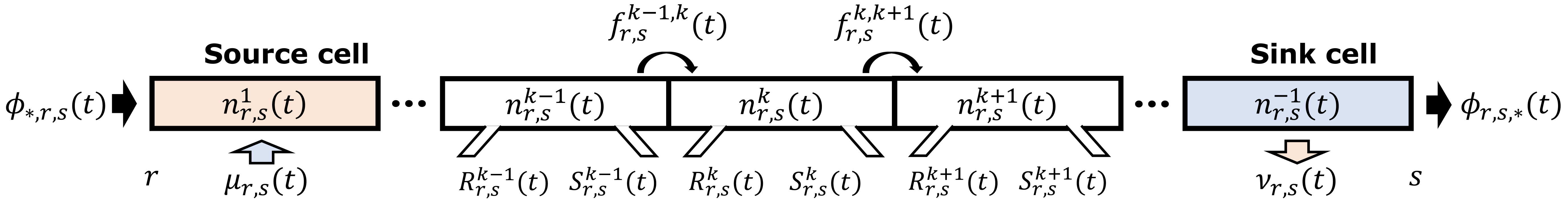}
    \caption{The illustration of the ACTM.}
    \label{fig:actm}
\end{figure}

The ACTM is a generalized variant of the Cell Transmission Model (CTM) \citep{CTM1,CTM2}. In the ACTM, each freeway is split into several cells which is the atomic unit of the traffic flow evolution. The difference between the ACTM and CTM is the way to model the ramps. 
Given a link $e_{r,s} \in \mathbf{E}, r, s \in \mathbf{V}$ representing the head and tail of the link, the length of road $e_{r,s}$ is $l_{r,s}$, and the length of the cell $\delta_{r,s}$ is computed as $\delta_{r,s} = v_{r,s}^{max} \cdot \Delta t$, where $v_{r,s}^{max}$ is the free-flow speed on road $e_{r,s}$ and $\Delta t$ is the unit time interval. The illustration of the ACTM is shown in Figure~\ref{fig:actm}, where $n_{r,s}^{k}(t)$ represents the vehicle number in cell $k$ on road $e_{r,s}$ at time $t$, $f_{r,s}^{k,k+1}$ is the internal traffic flow from cell $k$ to $k+1$ on road $e_{r,s}$ at time $t$, $R_{r,s}^{k}(t)$ denotes the on-ramp flow into cell $k$ on road $e_{r,s}$ at time $t$, $S_{r,s}^{k}(t)$ represents the off-ramp flow on road $e_{r,s}$ from cell $k$ at time $t$, $\mu_{r,s}(t)$ is the trip number started on road $e_{r,s}$ at time $t$, $\nu_{r,s}(t)$ is the trip number finished on road $e_{r,s}$ at time $t$, $\phi_{*,r,s}(t)$ is the external traffic flow from upstream roads into road $e_{r,s}$ at time $t$, and $\phi_{r,s,*}(t)$ is the external traffic flow from road $e_{r,s}$ to downstream roads at time $t$.

We name the first cell of the road $e_{r,s}$ as the ``source cell'' and the last cell of the road $e_{r,s}$ as the ``sink cell'', respectively. In the ACTM, the conservation law of traffic flow in source cells, sink cells,  general cells and on-ramp cells are formulated differently. For the conciseness of the paper, the conservation law in source cells, sink cells and general cells are discussed in \ref{apx:conservation_law}. Here, we only show the conservation law of traffic flow in on-ramp cells in Equation~\ref{eq:con_up}:
\begin{equation}
    \begin{aligned}
        n_{r',s'}^{-1}(t) &= n_{r',s'}^{-1}(t-1) + f_{r',s'}^{-2,-1}(t) - R_{r,s}^{k}(t), \\
        R_{r,s}^k(t) &= \min \left\{ n_{r',s'}^{-1}(t-1) + f_{r',s'}^{-2,-1}(t), \zeta \left(\hat{n}_{r,s} - n_{r,s}^{k}(t) \right), c_{r',s'}(t) \right\}, \\
        c_{r',s'}(t) &= \min \left\{ q_{r',s'}^{max}, q_{r,s}^{max} \right\}, \quad  e_{r,s}^{k} \text{ is not metered}, \\ 
        c_{r',s'}(t) &= \rho_{r,s}^{k}(t) \min \left\{ q_{r',s'}^{max}, q_{r,s}^{max} \right\}, \quad  e_{r,s}^{k} \text{ is metered}, \\
        &\qquad \qquad \forall r,s \in \mathbf{V}, \ 2 \leq k \leq \lceil l_{r,s} / \delta_{r,s} \rceil - 1,
    \end{aligned}
    \label{eq:con_up}
\end{equation}
\noindent where $n_{r',s'}^{-1}(t)$ is the number of vehicles in the last cell of the road $e_{r',s'}$, and  $e_{r',s'}$ is the on-ramp of  $e_{r,s}$. $f_{r,s}^{-2,-1}(t)$ is the internal traffic flow from the second to last cell to the last cell (sink cell) on road $e_{r',s'}$ at time $t$. $\zeta$ is on-ramp flow allocation coefficient indicating the available space on the mainline for the on-ramp flow. $\hat{n}_{r,s}$ is the jam density on road $e_{r,s}$. $c_{r',s'}(t)$ is the inflow capacity from the on-ramp to the mainline flow. If the ramp is not metered, the inflow capacity should be equal to the minimal volume in mainline and on-ramp capacity, $q_{r',s'}^{max}$ and $q_{r,s}^{max}$. If the ramp is metered, the inflow capacity is determined by the metering rate $\rho_{r,s}^{k}(t)$, on-ramp, and mainline capacity. Particularly, the metering rate $\rho_{r,s}^{k}(t)$ is controllable by the DRL.

The on-ramp flow should also satisfy the following conditions:
\begin{align}
    R_{r,s}^k(t) &\leq n_{r',s'}^{-1}(t-1) + f_{r',s'}^{-2,-1}(t) \label{eq:up_con1}, \\
    R_{r,s}^k(t) &\leq \zeta \left(\hat{n}_{r,s}(t) - n_{r,s}^{k}(t) \right) \label{eq:up_con2} ,\\
    R_{r,s}^k(t) &\leq c_{r',s'}(t) \label{eq:up_con3}.
\end{align}
where Equation~\ref{eq:up_con1} fulfills that the on-ramp flow should not surpass the demand,  Equation~\ref{eq:up_con2} prevents the mainstream from overload, and Equation~\ref{eq:up_con3} limits the on-ramp flow to the metered or non-metered capacity.

\subsubsection{Generalized bathtub model}
The mechanism of the bathtub model is different from that of the ACTM. In a generalized bathtub model, the movement of vehicles is no longer determined by the cell evolution but by the MFD, which models the relationship between space-mean traffic speed and regional vehicle accumulation. Hence, in a homogeneous region, all vehicles share the same speed at the same time. 

Mathematically, the conservation law in a region driven by the bathtub model is formulated in Equation~\ref{eq:con_bathtub}.
\begin{equation}
    \overline{N}_d(t) = \overline{N}_d(t-1) + \overline{\Mu}_d(t) + \overline{\Phi}_{*,d}(t) - \overline{\Nu}_d(t) - \overline{\Phi}_{d,*}(t), \quad d = 1, 2 \dots, \mathcal{N}_d,
    \label{eq:con_bathtub}
\end{equation}
\noindent where $\overline{N}_d(t)$ is the vehicle accumulation in region $d$ at time $t$. $\overline{\Mu}_d(t)$ is the number of trips begun at region $d$ at time $t$. $\overline{\Nu}_d(t)$ is the number of trips finished at region $d$ at time $t$. $ \overline{\Phi}_{*,d}(t)$ and $ \overline{\Phi}_{d,*}(t)$ is the number of inflow and outflow of region $d$ at time $t$, respectively.

In ACTM, the evolution of the traffic flow is actuated by moving vehicles between different cells. The relative location of a vehicle on the road can be derived based on the occupied cells. In the bathtub model, such a location is difficult to track since the vehicles are on different routes with different OD pairs. To tackle this problem, the quantity of the remaining distance is utilized in the bathtub model to indicate the relative location in a region, and to guide the movement of vehicles. In the beginning, when vehicles start trips or travel from other regions, the remaining distance may not be unified. However, when they finish trips or travel to other regions, the remaining distance should be equal to zero. Supposed that $N_{d}(t, \xi)$ represents the number of vehicles in region $d$ with a remaining distance of $\xi$ and time $t$, the evolution of vehicles modeled by the bathtub model is formulated in Equation~\ref{eq:evo_bathtub_non_term}, \ref{eq:evo_bathtub_term}, and \ref{eq:bathtub_mfd} as follows:
\begin{align}
    N(t, \xi) &= N(t-1, \xi + V_d(t) \Delta t) + \Phi_{*,d}(t, \xi) + \Mu_{d}(t, \xi), \quad \xi > 0,  d = 1, 2 \dots, \mathcal{N}_d \label{eq:evo_bathtub_non_term}, \\ 
    N(t, 0) &= N(t-1, V_d(t) \Delta t) - \overline{\Phi}_{d,*}(t) - \overline{\Nu}_{d}(t), \quad d = 1, 2 \dots, \mathcal{N}_d \label{eq:evo_bathtub_term}, \\ 
    V_d(t) &= \hat{V}\left(\frac{\overline{N}_{d}(t)}{L_{d}^{sum}}\right), \quad d = 1, 2 \dots, \mathcal{N}_d, \label{eq:bathtub_mfd}
\end{align}
\noindent where $V_d(t)$ represents the average speed in region $d$ at time $t$, $\hat{V}_d(\cdot)$ is the MFD of region $d$, $\Phi_{*,d}(t, \xi)$ is the vehicle inflow from upstream regions to region $d$ with the remaining distance of $\xi$ at time $t$, $\mathcal{U}_{d}(t, \xi)$ is the vehicle numbered started in region $d$ with a remaining distance of $\xi$ at time $t$, and $L_{d}^{sum}$ is the total length of roads in region $d$.

Equation~\ref{eq:evo_bathtub_non_term} models the evolution of vehicles with non-zero remaining distance, where only inflow and departing vehicles will show up in the region. Equation~\ref{eq:evo_bathtub_term} depicts the evolution of vehicles with zero remaining distance. If the remaining distance of a vehicle is zero, the vehicle will either be transferred to other regions or completes the trip. Equation~\ref{eq:bathtub_mfd} illustrates that the regional average speed is correlated with the average vehicle density in a region. 

The derivation of the external traffic flow is presented in Equation~\ref{eq:bathtub_ex_flow} for $d = 1, 2 \cdots, \mathcal{N}_d$.
\begin{equation}
    \begin{aligned}
        \overline{\Phi}_{*,d}(t) &=  \sum_{i \in \Psi^{-}(d)} \overline{\Phi}_{i, d}(t), \\
        \overline{\Phi}_{d,*}(t) &=  \sum_{j \in \Psi^{+}(d)} \overline{\Phi}_{d, j}(t), \\
        \overline{\Phi}_{i, d}(t) &= \hat{\Phi} \left( \left\{ \mathfrak{D}_{j}(t) \vert \forall j \in \Psi^{-}(d) \right\}, \left\{ \mathfrak{S}_{k}(t) \vert \forall k \in \Psi^{+}(d) \right\} \right), \\
        \overline{\Phi}_{d, j}(t) &= \hat{\Phi} \left( \left\{ \mathfrak{D}_{i}(t) \vert \forall i \in \Psi^{-}(j) \right\}, \left\{ \mathfrak{S}_{k}(t) \vert \forall k \in \Psi^{+}(j) \right\} \right), 
    \end{aligned}
    \label{eq:bathtub_ex_flow}
\end{equation}
\noindent where $\Phi_{i,d}(t)$ denotes the external traffic flow from region $i$ into region $d$. $\hat{\Phi}(\cdots)$ represents a function that allocates the external flows in different regions by transferring vehicles between regions. $\mathfrak{D}_{j}$ is the demand in region $d$ at time $t$, and $\mathfrak{S}_{d}$ is the supply in region $d$ at time $t$. $\Psi_{d}^{-}$ and $\Psi_{d}^{+}$ means sets of upstream and downstream regions that directly connect to region $d$, respectively. 

Perimeter control  aims to improve the network throughput by limiting the traffic inflow into congested urban areas at the boundary of regions. Hence, the quantitative value should be related to the maximal capacity at the boundary. Equation~\ref{eq:bathtub_demand_supply} shows the computation of the regional supply, demand, and boundary capacity.
\begin{equation}
    \begin{aligned}
    \mathfrak{D}_{d}(t) &= \min \left\{ N_{d}(t, 0) - \overline{\Nu}_{d}(t), C_{d,+}^{margin} \right\}, \\
    \mathfrak{S}_{d}(t) &= \min \left\{ \hat{N}_{d} - \overline{N}_{d}(t) - \Mu_{d}(t), C_{d,-}^{margin} \right\}, \\
    C_{d,+}^{margin} &= \sum_{e_{r,s} \in E, r \in \mathbf{V}_d, s \not\in \mathbf{V}_d} q_{r,s}^{max}, \\
    C_{d,-}^{margin} &= \begin{cases} 
    \sum_{e_{r,s} \in E, r \not\in \mathbf{V}_d, s \in \mathbf{V}_d} q_{r,s}^{max} & \mathbf{G}_d \text{ is not controlled} \\
    \Rho_{d}(t) \sum_{e_{r,s} \in E, r \not\in \mathbf{V}_d, s \in \mathbf{V}_d} q_{r,s}^{max} & \mathbf{G}_d \text{ is controlled} \\
    \end{cases}, \\
    \hat{N}_{d} &= \sum_{e_{r,s} \in E, r, s \in \mathbf{V}_{d}} \hat{n}_{r,s}, \\
    \end{aligned}
    \label{eq:bathtub_demand_supply}
\end{equation}
\noindent where $C_{d,+}^{margin}$ is the boundary capacity for vehicle outflow  while $C_{d,-}^{margin}$ is the boundary capacity for vehicle inflow, and $\Rho_{d}(t)$ is the control rate for region $d$ at time $t$. If the region $d$ is controlled, the vehicle inflow should not exceed the total road capacity at boundaries $\sum_{e_{r,s} \in E, r \not\in \mathbf{V}_d, s \in \mathbf{V}_d} q_{r,s}^{max}$ with a fraction of $\Rho_{d}(t)$. Let $\hat{N}_{d}$ represent the total jam density in region $d$. 
Additionally, the relationship between the flow evolution and conservation law in the bathtub model is shown in Equation~\ref{eq:bathtub_corr} for $d = 1, 2 \cdots, \mathcal{N}_d$.
\begin{equation}
    \begin{aligned}
        \overline{N}_{d}(t) &= \int_{\xi=0}^{L_{d}^{max}}N(t, \xi) d\xi, \\
        L_{d}^{sum} &= \sum_{ r, s \in \mathbf{V}_d} l_{r,s} ,\\
        L_{d}^{max} &\leq L_{d}^{sum}, 
    \end{aligned}
    \label{eq:bathtub_corr}
\end{equation}
\noindent where $L_{d}^{max}$ is the length of the longest route in region $d$.

\subsection{Demonstration-guided DRL method}
The proposed DRL method contains two modules, the decision-making module and the demonstrator. The decision-making module aims to learn an optimal control policy by exploring the environment, and the demonstrator generates demonstration data to guide the learning of the decision-making module. 

\subsubsection{Partially Observable Markov Decision Process}
The interaction between agents and the environment can be formulated as a Partially Observable Markov Decision Processes (POMDPs), in which agents can access limited information from the environment. The POMDPs can be depicted by a 6-tuple $(\mathcal{X}, \mathcal{U}, \mathcal{P}, \mathcal{R}, \Omega, \mathcal{O})$. At each timestep $t$, a state incorporating the full environment information is $\mathbf{x}_t \in \mathcal{X}$. An agent can observe the environment and obtain the observation $\mathbf{o}_t \in \Omega$ with a probability of $\mathbf{o} \sim \mathcal{O}(x)$. The environment will evolve  into a new state $\mathbf{x}_{t+1} \sim P(\mathbf{x}_t, \mathbf{u}_t)$ when the agent takes an action $\mathbf{u}_t \in \mathcal{U}$. The objective of a single agent at time $t$ is to maximize the discounted cumulative rewards it will obtain in the future before the termination of the interaction, which is formulated in Equation~\ref{eq:rl_goal}.
\begin{equation}
    G_t = \sum_{\tau=t}^{\hat{T}^{term}} \lambda^{\tau - t} \mathbf{r}_{\tau},
    \label{eq:rl_goal}
\end{equation}
\noindent where $G_t$ represents the discounted cumulative reward from the start time $t$, $\hat{T}^{term}$ is the termination step, and $\lambda$ is the discount factor that balances the short-term and long-term returns. The action value function $Q(\mathbf{x}_t, \mathbf{u}_t)$ measures the expectation of reward when an agent takes action $\mathbf{u}_t$ at state $\mathbf{x}_t$ at time $t$, which is defined in Equation~\ref{eq:action_value}:
\begin{equation}
    Q_\pi(\mathbf{x}_t, \mathbf{u}_t) = \mathbb{E}\left( G_{t} \vert \mathbf{x}_t, \mathbf{u}_t  \right).
    \label{eq:action_value}
\end{equation}

Specifically, in this paper, the definitions of an agent, state, observation, action, and reward are shown as follows.

\paragraph{\textbf{Agent}}
An agent in the meso-macro network could be either a ramp or a perimeter. While the former type of agent controls the inflow from ramps, the latter type of agent controls the inflow into congested urban areas. Since this study models traffic control as a POMDP problem, an agent can  obtain partial observations and historical information $\mathbf{o}_t$ and $\mathbf{h}_t$ rather than state information $\mathbf{x}_t$. The agent takes an action $\mathbf{u}_t$ from its own policy $\pi(\mathbf{o}_t, \mathbf{h}_t)$, and receives a reward $\mathbf{r}_t$ from the environment.

\begin{figure}[h]
  \centering    
  \subfigure[An example of an ACTM agent.] {
   \label{fig:obs_actm}     
  \includegraphics[width=0.4\columnwidth]{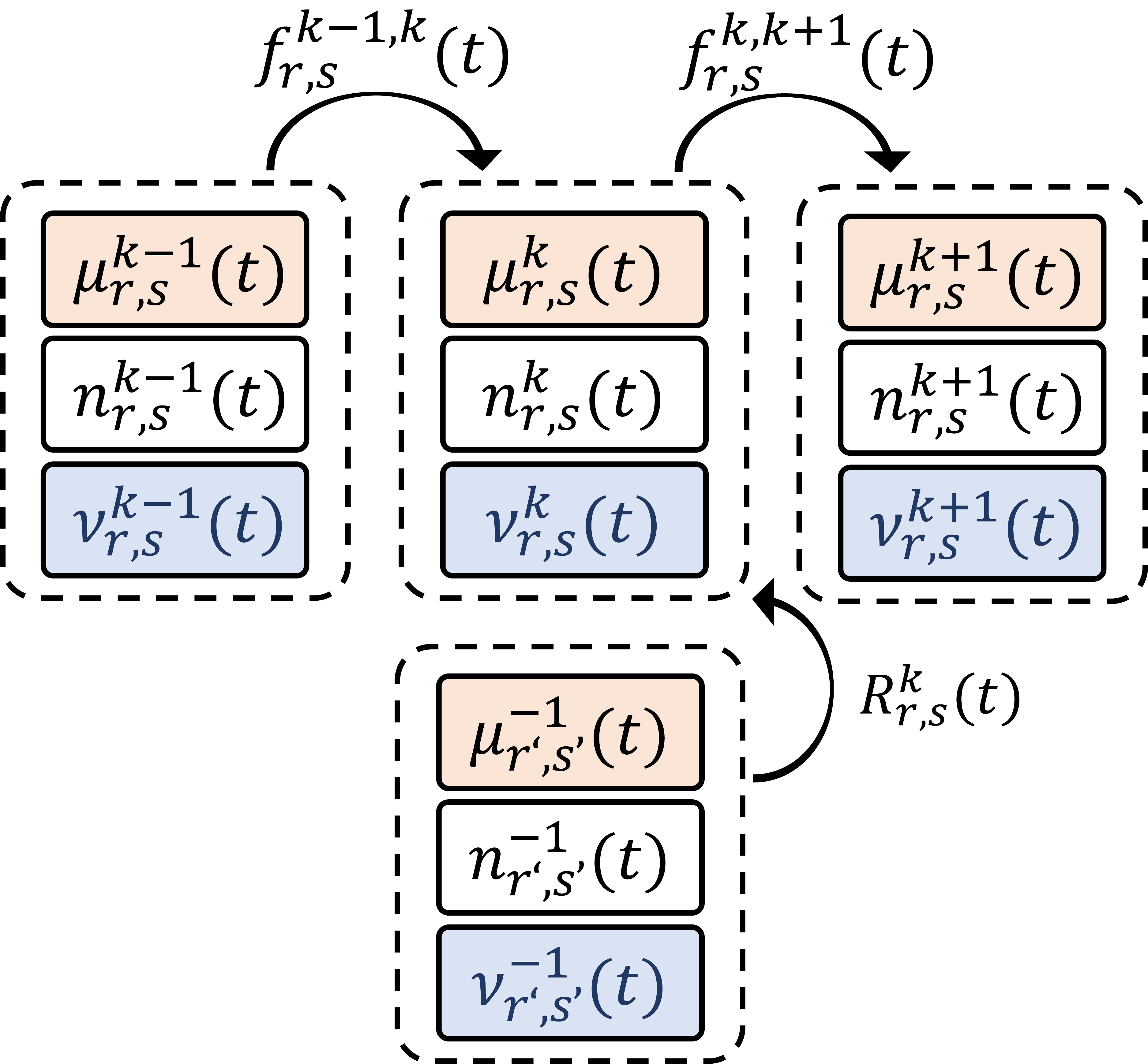}
  }
  \subfigure[An example of a bathtub agent.] { 
  \label{fig:obs_bathtub}     
  \includegraphics[width=0.4\columnwidth]{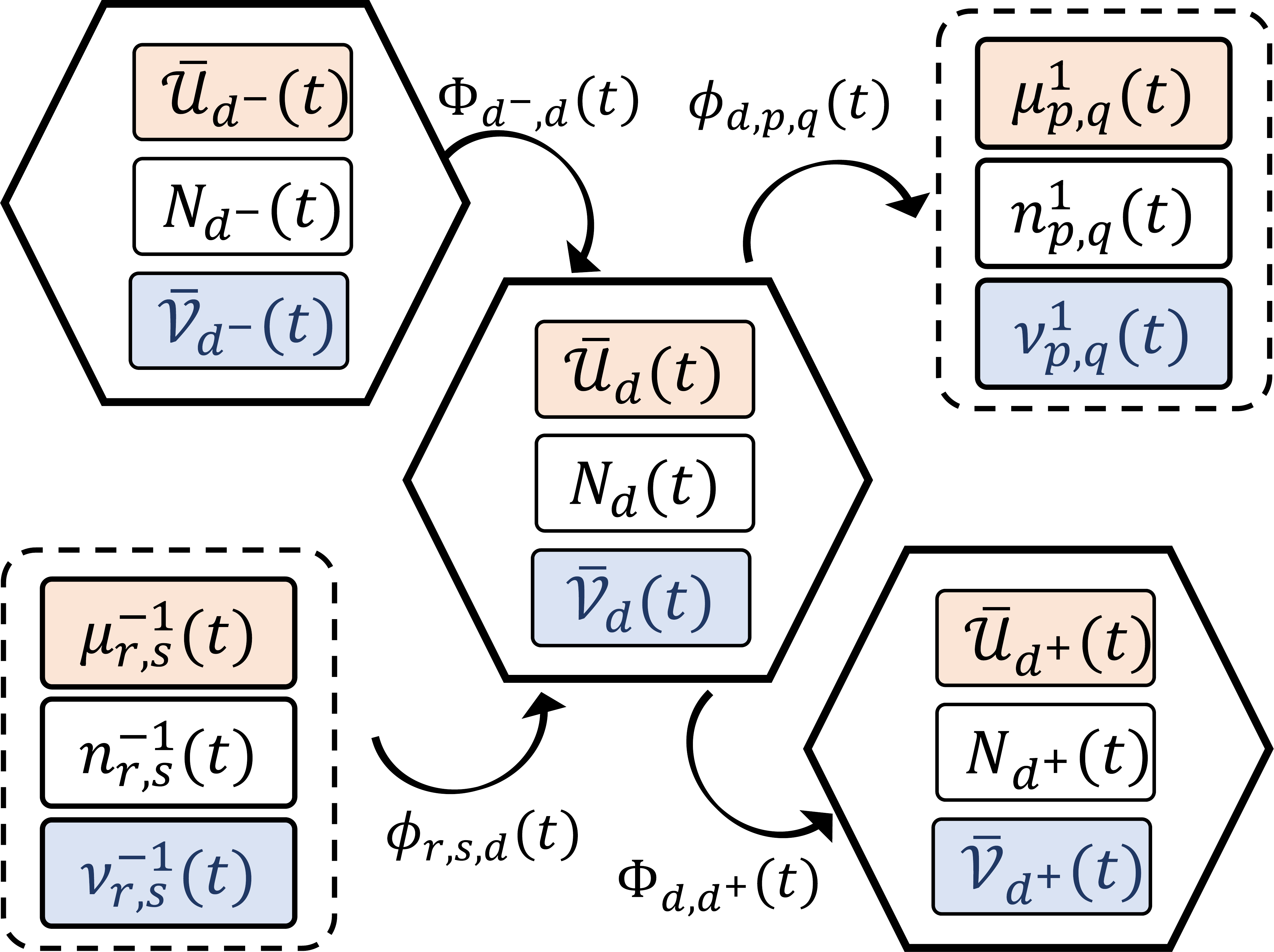}
  }
  \label{fig:obs_agents}
  \caption{Two example agents of the ACTM and bathtub region.}
\end{figure}

\paragraph{\textbf{State and observation}}
In the road network, the state information consists of the trip source, sink, vehicle accumulation, and traffic flow in the ACTM cells and the generalized bathtub regions for all the roads and regions. However, in the real world, an agent may not receive all state information in the whole network due to the limitation of network bandwidth and information processing speed. A plausible setting would be that agents can  receive local information from their neighbors (\textit{i.e.,} adjacent ACTM cells and bathtub regions). The local information consists of node information and edge information as follows:
\begin{itemize}
    \item For an ACTM cell, the node information includes the trip started number (if it is a source cell), the trip completed number (if it is a sink cell), and the vehicle number in the ego cell. The edge information includes the internal traffic flow between different cells. Then for a ramp, the observation consists of local information of adjacent cells. An example of the observation of a ramp cell is shown in Figure~\ref{fig:obs_actm}. The adjacent cells include the upstream cell, the downstream cell, and the connected ramp cell. There are 12 variables in node information, and 3 variable in edge information. In total, the dimension of observation of a ramp agent is 15.
    \item  For a bathtub region, the node information includes the trip started number, trip completed number, and the vehicle accumulation in this region, while the edge information incorporates the external traffic flow from or to different ACTM cells or bathtub regions. The observation consists of local information of adjacent cells and regions. An example of a perimeter cell is shown in Figure~\ref{fig:obs_bathtub}. The neighbors include source cells and sink cells in the ACTM, and regions in the bathtub model. Since the number of surroundings is not fixed for a perimeter agent, the dimension varies in different perimeter agents.
\end{itemize}

\paragraph{\textbf{Action}}
In a ramp metering task, the metering rate $\rho_{r,s}^{k}(t)$, which is a continuous scalar ranging in $[\mathbf{u}_{min}, \mathbf{u}_{max}]$ in Equation~\ref{eq:con_up} is set to adjust the percentage of vehicle inflow from on-ramps. In the perimeter control task, the boundary inflow rate $\Rho_{d}(t)$, which is also a continuous scalar ranging in $[\mathbf{U}_{min}, \mathbf{U}_{max}]$ in Equation~\ref{eq:bathtub_demand_supply} is defined in each region to control the inflow vehicles from other regions to ease congestion. Since it is challenging to achieve a continuous coordinated control for multiple ramps and perimeters simultaneously due to the huge action space when the number of controllable agents is large, we cast the continuous action space into a discrete decision-making process. There are three options for each agent, raise the control rate by a minimal change step $\Delta \mathbf{u}$, keep the current control rate and reduce the control rate by $\Delta \mathbf{u}$. Hence the action space for each agent falls in $\mathbf{u}_t \in \mathbb{R}^{3}$.

\paragraph{\textbf{Reward}}
The coordinated ramp metering and perimeter control framework aims to minimize the TTT of all travelers in the network. Mathematically, the minimization of the average TTT for all travelers is expressed in Equation~\ref{eq:control_goal}, where $\mathcal{N}_p$ is the number of trips, and $T_{i}^{end}$ and $T_{i}^{start}$ is the end time and start time for trip $i$. 
\begin{equation}
    \min \frac{1}{\mathcal{N}_p} \sum_{i=0}^{\mathcal{N}_p} \left( T_{i}^{end}  - T_{i}^{start}\right)
    \label{eq:control_goal}
\end{equation}

If we integrate Equation~\ref{eq:control_goal} along with the timeline, the minimization of the average TTT can be approximated to the maximization of the sum of vehicle completion from time $t$ to $t + \Delta t$. The derivation is shown in Equation~\ref{eq:control_goal_trans}.
\begin{equation}
    \begin{aligned}
        \min \frac{1}{\mathcal{N}_p} \sum_{i=0}^{\mathcal{N}_p} \left( T_{i}^{end}  - T_{i}^{start}\right) &\approx \max \frac{\Delta t}{\mathcal{N}_p} \sum_{i=0}^{\lceil T^{term} / \Delta t \rceil } \left( \mathcal{N}_{inj}(i\Delta t) - \mathcal{N}_{run}(i\Delta t) \right), \\
        &= \max \frac{\Delta t}{\mathcal{N}_p} \sum_{i=0}^{\lceil T^{term} / \Delta t \rceil } \mathcal{N}_{com} \left( i \Delta t, (i+1) \Delta t \right), \\
        &  \leftrightarrow \sum_{i=0}^{\lceil T^{term} / \Delta t \rceil } \max \mathcal{N}_{com} \left( i \Delta t, (i+1) \Delta t \right), \\
    \end{aligned}
    \label{eq:control_goal_trans}
\end{equation}
\noindent where $\mathcal{N}_{inj}(i \Delta t)$ denotes the accumulative injected vehicle at time $i \Delta t$, $\mathcal{N}_{run}(i \Delta t)$ represents the number of running vehicles at time $i \Delta t$, and $\mathcal{N}_{com}(i \Delta t, (i+1) \Delta t)$ represents the number of completed vehicles from time $i \Delta t$ to time $(i+1) \Delta t$. The optimal solution is lower bounded by $ \max \sum_{i=0}^{\lceil T^{term} / \Delta t \rceil } \mathcal{N}_{com} \left( i \Delta t, (i+1) \Delta t \right) \geq  \sum_{i=0}^{\lceil T^{term} / \Delta t \rceil } \max \mathcal{N}_{com} \left( i \Delta t, (i+1) \Delta t \right)$, which means that agents can greedily take actions that can maximize the vehicle completion number for now. Though it may not be the optimal solution for the minimal TTT, the network efficiency can also be improved by taking the control policy that makes most trips completed in the current time interval. Therefore, the step-wise rewards for all controllable agents are set as the number of completed trips from $t$ to $t + \Delta t$. Furthermore, we normalize the reward with a baseline constant $C_r$. The final reward is defined in Equation~\ref{eq:rl_reward}.
\begin{equation}
    \mathbf{r}_t = \mathcal{N}_{com}(t, t + \Delta t) - C_r
    \label{eq:rl_reward}
\end{equation}

\subsubsection{Deep Q network}
In the fully observable MDP for the model-free control problem, the action value function $Q(\mathbf{x}, \mathbf{u})$ is utilized to evaluate the expectation of return after taking an action $\mathbf{u}$ at state $\mathbf{x}$ under the policy $\pi$. In simplified control problems with finite states, the $Q(\mathbf{x}, \mathbf{u})$ can be derived from tabular updating methods (\textit{e.g.}, Q-Learning \citep{qlearning}, sarsa \citep{sarsa}, \textit{etc.}). The policy is taking the action with the maximal $Q$ value, which is formulated as 
\begin{equation}
    \pi \left(\mathbf{u}_t \vert \mathbf{x}_t \right) = \begin{cases} 1, \quad \text{if } Q(\mathbf{x}_t, \mathbf{u}_t) = \max_{ \hat{\mathbf{u}}_t} \left( Q \left( \mathbf{x}_t, \hat{\mathbf{u}_t} \right) \right) \\
    0  \end{cases}.
    \label{eq:pi}
\end{equation} For a complex control problem, the $Q(\mathbf{x}, \mathbf{u})$ is difficult to be acquired from direct tabular searching since the $Q(\mathbf{x}, \mathbf{u})$ is hard to converge in a limited time with a large state and action space. To solve $Q(\mathbf{x}, \mathbf{u})$ in complex control problems, the action value function approximation, which treats the $Q(\mathbf{x}, \mathbf{u})$ as a function of state and action $\hat{Q}: \mathbf{x} \times \mathbf{u} \rightarrow Q \in \mathbb{R}$, is leveraged to approximate the real action value $Q(\mathbf{x}, \mathbf{u})$. 
Two significant contributions have been made to stabilize the training process, the experience replay buffer and the target network. The replay buffer collects transition tuples of state, action, reward and next state $(\mathbf{x}_t, \mathbf{u}_t, \mathbf{r}_t, \mathbf{x}_t')$ for each time $t$ in a sequential manner. The DQN model can request the transition in the buffer replay at the training stage in multiple times, which enhances the sample efficiency. Furthermore, it breaks down the temporal correlation between transitions, which may enhance the final control performance. The usage of the target network leads the Q-network (source network) to converge by cloning the weights of the Q-network periodically. To optimize the Q-network weight, the loss function is defined as 
\begin{equation}
    L_{RL} = \left( \mathbf{r}_t + \lambda \max_{\mathbf{u}_{t+1}} Q(\mathbf{x}_{t+1}, \mathbf{u}_{t+1}; \theta^{-}) -  Q(\mathbf{x}_{t}, \mathbf{u}_{t}; \theta) \right)^2,
\end{equation}
\noindent where $\theta^{-}$ is the weight of the target network and $\theta$ is the weight of the Q network. 

\subsubsection{Demonstration-guided DQN}
To enhance the performance of coordinated control of ramps and perimeters, we incorporate demonstrators to guide agents in learning control policies. There are three rules when selecting demonstrators. 1) The demonstrator should be simple to acquire; 2) The demonstrator should be efficient; 3) The performance of the demonstrator should be decent. Based on these rules, we choose ALINEA \citep{alinea} as the demonstrator for the ramp agent and Gating \citep{pi1_pm} as the demonstrator for the perimeter agent, respectively. More information regarding the demonstrators can be referred to \ref{apx:demonstrator}.

The cross entropy can be utilized to measure the discrepancy between a demonstrator policy $\overline{\pi} \left({\mathbf{u}_t} \vert \mathbf{x}_{t} \right)$, and the model policy $\pi \left({\mathbf{u}_t} \vert \mathbf{x}_{t} \right)$. In the proposed DRL method, the cross entropy is defined as:

\begin{equation}
    L_{CE}(t) = - \sum_{\mathbf{u}_{t} \in \mathbf{U}} \pi \left({\mathbf{u}_t} \vert \mathbf{x}_{t} \right) \log \overline{\pi} \left({\mathbf{u}_t} \vert \mathbf{x}_{t} \right).
    \label{eq:ce_org}
\end{equation}

The DRL method approximates the action value function $\hat{Q}: \mathbf{x} \times \mathbf{u} \rightarrow Q \in \mathbb{R}$, which takes the state and action as inputs and outputs the Q value. According to Equation~\ref{eq:pi}, the final action is selected as the one with the maximal Q value (\text{i.e., $\argmax_{\mathbf{u}_t} Q \left(\mathbf{x}_t, \mathbf{u}_t \right)$}). As the ``argmax'' operation cannot contribute gradient to the DRL method in the backpropagation, we involve a dummy policy $\tilde{\pi} \left({\mathbf{u}_t} \vert \mathbf{x}_{t}\right)$, which takes the ``softmax'' operation of the original Q values, shown in Equation~\ref{eq:dummy_policy}.
\begin{equation}
    \tilde{\pi} \left({\mathbf{u}_t} \vert \mathbf{x}_{t}\right) = \frac{\exp{\left(Q \left({\mathbf{u}_t }, \mathbf{x}_t; \theta \right) \right)}}{\sum_{\hat{\mathbf{u}}_{t} \in \mathbf{U}} \exp{\left(Q \left({\hat{\mathbf{u}}_t}, \mathbf{x}_t; \theta \right)\right)}}
    \label{eq:dummy_policy}
\end{equation}

The demonstration loss that distinguishes the demonstrator and the DRL method is represented in Equation~\ref{eq:demonstration_loss}
\begin{equation}
    L_{D}(t) = - \sum_{\mathbf{u}_{t} \in \mathbf{U}} \tilde{\pi} \left({\mathbf{u}_t} \vert \mathbf{x}_{t} \right) \log \overline{\pi} \left({\mathbf{u}_t} \vert \mathbf{x}_{t} \right)
    \label{eq:demonstration_loss}
\end{equation}

Replacing $\pi \left({\mathbf{u}_t} \vert \mathbf{x}_{t} \right)$ by $\tilde{\pi} \left({\mathbf{u}_t} \vert \mathbf{x}_{t} \right)$, we are able to calculate the gradient and conduct backpropagation to minimize the demonstration loss $L_{D}(t)$ in Equation~\ref{eq:demonstration_loss}. 
Note that $\tilde{\pi} \left({\mathbf{u}_t} \vert \mathbf{x}_{t} \right)$ does not represent the actual policy for taking actions. In contrast, it approximates the actual policy and ensures the easy derivation of gradients when training the DRL.


\subsubsection{Solution framework}
\label{sec:algorithm}

\begin{figure}[h]
    \centering
    \includegraphics[width=1\textwidth]{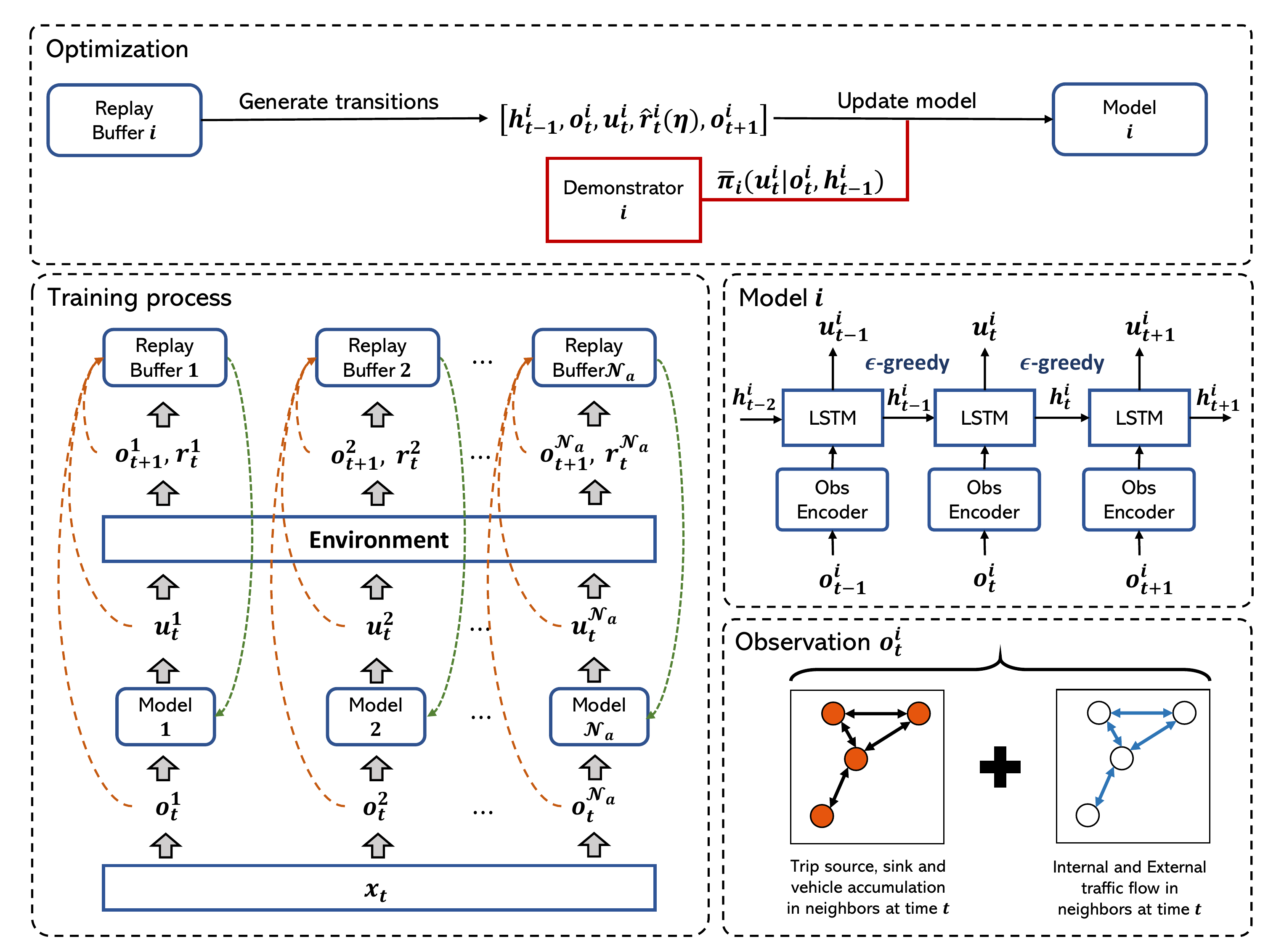}
    \caption{The framework of the proposed demonstrated guided DRL method.}
    \label{fig:model_framwork}
\end{figure}

Since we model the coordinated control problem as a POMDP, the historical information can be utilized to compensate for the deficiency in the state information. The action value function can be modeled as a function of observation, historical information, and action $\hat{Q}: \mathbf{o} \times \mathbf{h} \times \mathbf{u} \rightarrow Q \in \mathbb{R}$. The framework of the proposed demonstration-guided DRL method is shown in Figure~\ref{fig:model_framwork}. For each agent $i$ at time $t$, it observes trip start, trip end, vehicle accumulation, and traffic flow information in neighboring roads and regions, $\mathbf{o}_{t}^{i}$. The observed information was initially encoded by the ``observation encoder'' module, which is a Multilayer Perceptron (MLP). The encoded information was combined with encoded historical information $\mathbf{h}_{t-1}^{i}$ in the Long short-term memory (LSTM) module which outputs the next historical information $\mathbf{h}_{t}^{i}$ and estimated action value $Q_i\left(\mathbf{o}_t^i, \mathbf{h}_{t-1}^{i}, \mathbf{u}_t^i\right)$ for all eligible actions. The output action is selected according to the $\epsilon$-greedy policy to balance the exploration of the environment and the exploitation of the model. Specifically, the action will be selected with the maximal action value, $\argmax_{\mathbf{u}_t} Q_i\left(\mathbf{o}_t^i, \mathbf{h}_{t-1}^{i}, \mathbf{u}_t^i\right)$, with the probability of $1 - \epsilon$, and will be selected randomly in the action space with the probability of $\epsilon$. Interacting with the environment, each agent $i$ will receive a reward $\mathbf{r}_t^i$ and it can observe the environment again to acquire knowledge $\mathbf{o}_{t+1}^i$, and the transition information includes a tuple of $\left[ \mathbf{h}_{t-1}, \mathbf{o}_t, \mathbf{u}_t, \mathbf{r}_t, \mathbf{o}_{t+1} \right]$.

To optimize the proposed DRL method, the final loss function for each agent is formulated in Equation~\ref{eq:final_loss_1} and ~\ref{eq:final_loss_2}. It incorporates two parts. The first part is the traditional DQN loss modeled in POMDP. The multi-step return mechanism that exposes further information to the current state is also incorporated to fasten the convergence of the DRL method. The second part is the demonstration loss which measures the discrepancy between the demonstrator policy and the agent policy.
\begin{equation}
    L^{i}(t) = L_{RL}^{i}(t) + \alpha_t L_{D}^{i}(t)
    \label{eq:final_loss_1}
\end{equation}

\begin{equation}
    \begin{aligned}
        L_{RL}^{i}(t) &= \left( \hat{\mathbf{r}}_t^{i}(\eta) +  \lambda^{\eta} \max_{\mathbf{u}_{t+1}^{i}} Q_i \left( \mathbf{o}_{t+1}^{i}, \mathbf{h}_{t}^{i}, \mathbf{u}_{t+1}^{i}; \theta_{i}^{-} \right) -  Q_i \left( \mathbf{o}_{t}^{i}, \mathbf{h}_{t-1}^{i}, \mathbf{u}_{t}^{i}; \theta_{i} \right) \right)^2 \\
        L_{D}(t) &= - \sum_{\mathbf{u}_{t} \in \mathbf{U}} \tilde{\pi}_i \left({\mathbf{u}_t^i} \vert \mathbf{o}_{t}^i, \mathbf{h}_{t-1}^i \right) \log \overline{\pi}_i \left({\mathbf{u}_t^i} \vert \mathbf{o}_{t}^i, \mathbf{h}_{t-1}^i \right) \\
        \hat{\mathbf{r}}_t(\eta) &= \sum_{i=1}^{\eta} \lambda^{\eta} \mathbf{r}_{t+i}, \\
    \end{aligned}
    \label{eq:final_loss_2}
\end{equation}

To balance the traditional DQN loss and demonstration loss, there is also a time-varying annealing factor $\alpha_t$, which is defined in Equation~\ref{eq:anneal}. There are 4 parameters that adjust the annealing factor in each epoch: $\alpha_{min}$, $\alpha_{step}$, $\alpha_{term}$, and $\alpha_{amp}$. In the training stage, the annealing factor is a large number at the beginning, hence $L_{RL}^{i}(t) \ll \alpha_t L_{D}^{i}(t)$. The DRL method is constrained by the demonstrator without any exploration to resemble supervised training. Such a constraint anneals with time, and when $\alpha_t$ is close to $0$, $L_{RL}^{i}(t) \gg \alpha_t L_{D}^{i}(t)$, the training loss is close to the DQN loss, which permits the agent to learn a better policy through exploration and exploitation of the environment. 

\begin{equation}
\begin{aligned}
    \Alpha(t) &= 1 - \frac{1}{1 + \exp(-\alpha_{min} - \alpha_{step} * t)} \\
    \alpha_t &= \begin{cases}
    \alpha_{amp} * \frac{\Alpha(t) - \Alpha(\alpha_{term})}{\Alpha(0) - \Alpha(\alpha_{term})} & , t \leq \alpha_{term} \\
    0 &, t > \alpha_{term}
    \end{cases}
\end{aligned}
\label{eq:anneal}
\end{equation}

The pseudo-code for learning the proposed demonstration-guided DRL method is presented in Algorithm~\ref{alg:model}. 

\begin{algorithm}
\SetKw{Agent}{agent}
\caption{Demonstration-guided DRL method for coordinated ramp metering and perimeter control }
\label{alg:model}
\For{\textnormal{Agent} $ i = 1, \dots, \mathcal{N}_a$}{
    Randomly initialize replay buffer $B^i$ to capacity $\mathcal{N}_B$\;
    Initialize the historical information $h^i_0 = \overrightarrow{0}$\;
    Initialize the Q-network with random weights $\theta_i$\;
    Clone the Q-network weight $\theta_i$ to the target network weight $\theta_i^-$ \;
}
\For{\textnormal{Epoch} $m = 1, \dots, \mathcal{N}_m$}{
    \For{\textnormal{Time Interval} $t = 1, \dots, T$}{
        \For{\textnormal{Agent} $ i = 1, \dots, \mathcal{N}_a$}{
            Select a random action $\mathbf{u}_t^i$ with probability of $\epsilon(m)$ \;
            Otherwise select $\mathbf{u}_t^i = \argmax_{\mathbf{u}} Q \left(\mathbf{o}_{t}^{i}, \mathbf{h}_{t-1}^{i}, \mathbf{u}_{t}^{i}; \theta_{i} \right)$ with probability of $1-\epsilon(m)$\;
            Obtain historical information $\mathbf{h}_t^{i}$;\
        }
        Execute actions $\left\{\mathbf{u}_t^1, \mathbf{u}_t^r, \dots, \mathbf{u}_t^{\mathcal{N}_a}\right\}$ in the simulator. Obtain rewards $\left\{\mathbf{r}_t^1, \mathbf{r}_t^2, \dots, \mathbf{u}_t^{\mathcal{N}_a}\right\}$ and observe $\left\{\mathbf{o}_{t+1}^1, \mathbf{o}_{t+1}^2, \dots, \mathbf{o}_{t+1}^{\mathcal{N}_a}\right\}$\;
        \For{\textnormal{Agent} $ i = 1, \dots, \mathcal{N}_a$}{
            Store transition $\left[ \mathbf{h}_{t-1}^i, \mathbf{o}_t^i, \mathbf{u}_t^i, \mathbf{r}_t^i, \mathbf{o}_{t+1}^i \right]$ in $B^i$\;
        }
    }
    \For{\textnormal{Optimization Step} $ k = 1, \dots, \mathcal{N}_o$}{
            \For{\textnormal{Agent} $ i = 1, \dots, \mathcal{N}_{a}$}{
                Sample random mini-batch of transitions $\left[ \mathbf{h}_{j-1}^i, \mathbf{o}_j^i, \mathbf{u}_j^i, \mathbf{r}_j^i, \mathbf{o}_{j+1}^i \right]$\;
                Update the $\theta_i$ by minimizing the loss in Equation~\ref{eq:final_loss_1}\;
            }
        }
    Update the annealing factor $\alpha_m$ according to Equation~\ref{eq:anneal}\;
    \If{$(m\mod \mathcal{N}_{c}) == 0$}{
        \For{\textnormal{Agent} $ i = 1, \dots, \mathcal{N}_{a}$}{
            Clone the Q-network weight $\theta_i$ to the target network weight $\theta_i^-$\;
            }
        }
}
 \end{algorithm}

\clearpage
\section{Numerical experiments}
\label{sec:experiments}
In this section, two case studies are presented to evaluate the proposed demonstration-guided DRL method for coordinated ramp metering and perimeter control. In the first case study, the effectiveness of the proposed method is tested in a small-scale network with 4 on-ramps and 4 perimeters. In the second case study, we examine the proposed framework in a real-world large-scale network in Kowloon District, Hong Kong SAR, with a real-world demand in the morning peak.

\subsection{Settings of experiments}
For the meso-macro network model, the unit time interval is set to one second, and agents will interact with the environment every 30 seconds.
The on-ramp flow parameter, $\gamma$, is set to 1 and the on-ramp flow allocation coefficient $\zeta$ is fixed to 1. 
The lower bound and the upper bound of the control rate for ramps and perimeters, $\mathbf{u}_{min}$,  $\mathbf{U}_{min}$, and $\mathbf{u}_{max}$, $\mathbf{U}_{max}$ are set to $0.1$ and $1$, respectively. 
The change step of the control rate $\Delta u$ is set to $0.05$, which means the current control rate will increase or decrease by $0.05$. Moreover, if the current control rate exceeds the upper bound or lower bound $\mathbf{u}_{min}$,  $\mathbf{U}_{min}$ and $\mathbf{u}_{max}$, $\mathbf{U}_{max}$, the actual control rate will be truncated. In real world, the daily traffic demand may fluctuate, resulting in uncertainty of the congestion patterns for ramps and perimeters. To replicate the randomness of the traffic demand, given a demand profile, the actual traffic demand follows a normal distribution where the mean is the given demand for each OD pair, and the standard deviation is 30\% of the demand mean.

For the DRL method, in both case studies, the number of neurons in a 3-layer MLP is 100, 100, and 3 respectively. The hyperparameters regarding the DRL method are the same except for the number of epochs. The choices of hyperparameters are listed in Table~\ref{tab:hyperparameter}. The value of $\epsilon(t)$ will decay exponentially in the training stage to trade off between exploration and exploitation. The Adam optimizer is chosen in both case studies for optimizing the DRL method with the learning rate of $3e^{-6}$. The batch size for each sampling in the replay buffer is 128. 
The replay buffer makes use of the queue with the First In First Out (FIFO) property. 

\begin{table}[h]
    \centering
    \begin{tabular}{p{0.20\columnwidth}|p{0.20\columnwidth}|p{0.5\columnwidth}}
        \hline
        \textbf{Hyperparameter} & \textbf{Value} & \textbf{Description}  \\
         \hline
        $\mathcal{N}_{B}$ & 30,000 & The size of the replay buffer for each agent. \\
        $\mathcal{N}_{m}$ & 100; 200 & The number of epochs to train all agents. (100 in case study \#1 and 200 in case study \#2) \\
        $\mathcal{N}_{o}$ & 1 & The period for training DRL methods. \\
        $\epsilon_{start}$ & 0.1 & The initial value for the $\epsilon$. \\
        $\epsilon_{end}$ & 0.01 & The final value for the $\epsilon$. \\
        $\epsilon_{last}$ & 100 & The duration of $\epsilon$ decayed in the training process. \\
        $\alpha_{min}$ & -3 & The minimal value used to normalize the value of $\alpha_t$. \\
        $\alpha_{step}$ & 0.12 & The normalization term for the time $t$ in the calculation of $\alpha_t$. \\
        $\alpha_{term}$ & 50 & The termination steps in the calculation of $\alpha_t$. \\
        $\alpha_{amp}$ & 200 & The amplification factor. \\
        $\eta$ & 30 & The length of multi-step returns. \\
        $\lambda$ & 0.99 & The discount factor for the cumulative rewards. \\
        batch size & 128 & \\
        learning rate & 3e-6 & \\
         \hline
    \end{tabular}
    \caption{The hyperparameters chosen in the case study \#1 and case study \#2.}
    \label{tab:hyperparameter}
\end{table}

All experiments are conducted on a Cloud server with Intel(R) Xeon(R) Platinum 8372HC CPU @ 3.40GHz, 2666MHz $\times$ 64GB RAM, 500GB SSD. While the proposed method takes about 8 hours for training in case study \#1, it takes about 5 days to train on the Hong Kong network.

\subsection{Case study \#1: A small network}

\begin{figure}[h]
    \centering
    \includegraphics[width=1\textwidth]{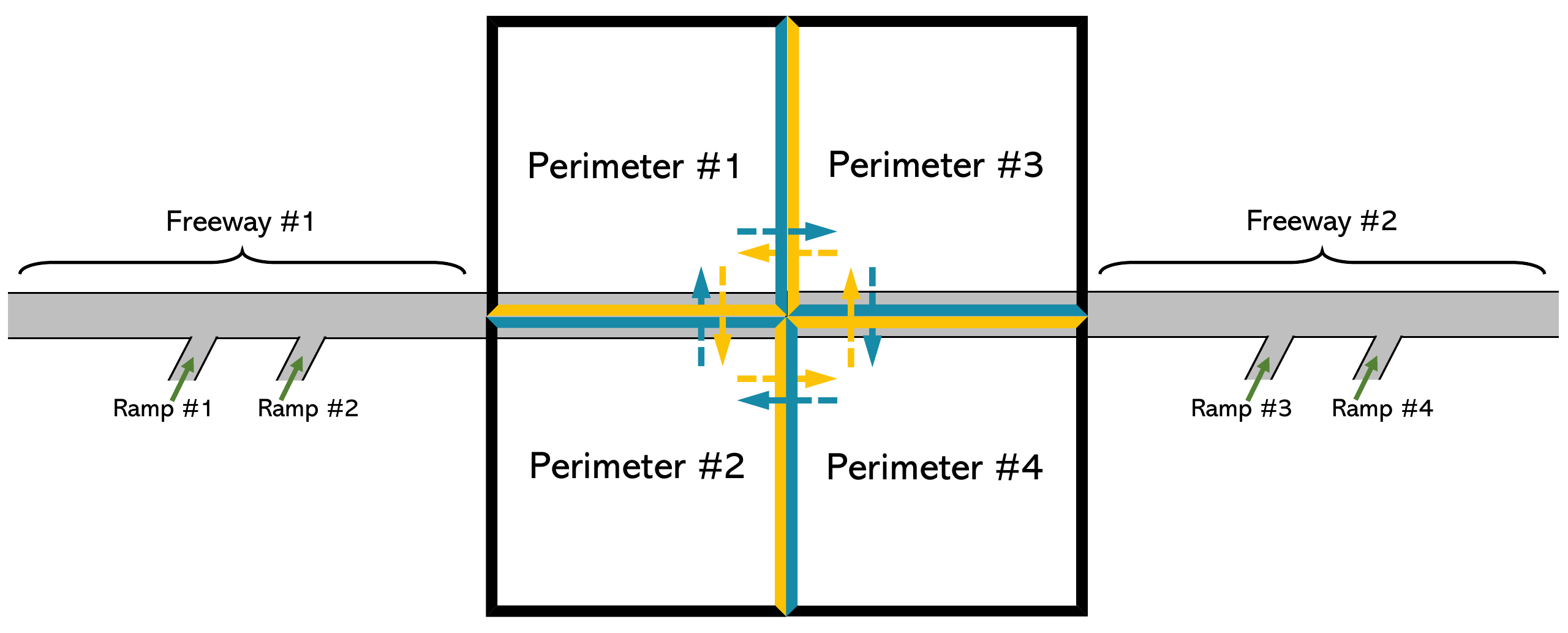}
    \caption{The topology sketch of the small network in case study \#1 (4 on-ramps and 4 perimeters).}
    \label{fig:study01_network}
\end{figure}

\begin{figure}[h]
  \centering    
  \subfigure[The sketch of the MFD for all bathtub models.] {
   \label{fig:mfd01}     
  \includegraphics[width=0.45\columnwidth]{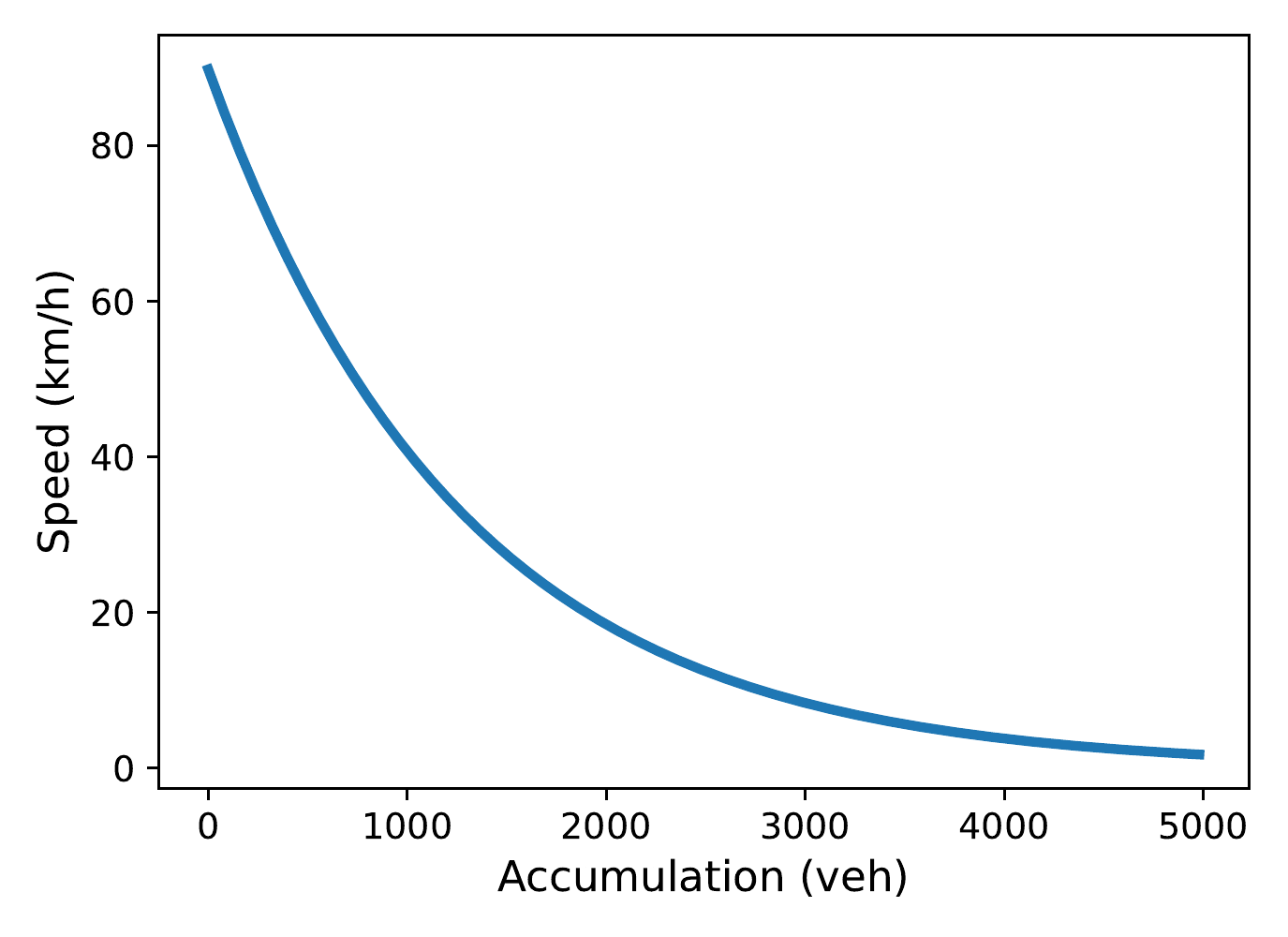}
  }
  \subfigure[The ratio of travel demand for each OD pair.] { 
  \label{fig:demand01}     
  \includegraphics[width=0.45\columnwidth]{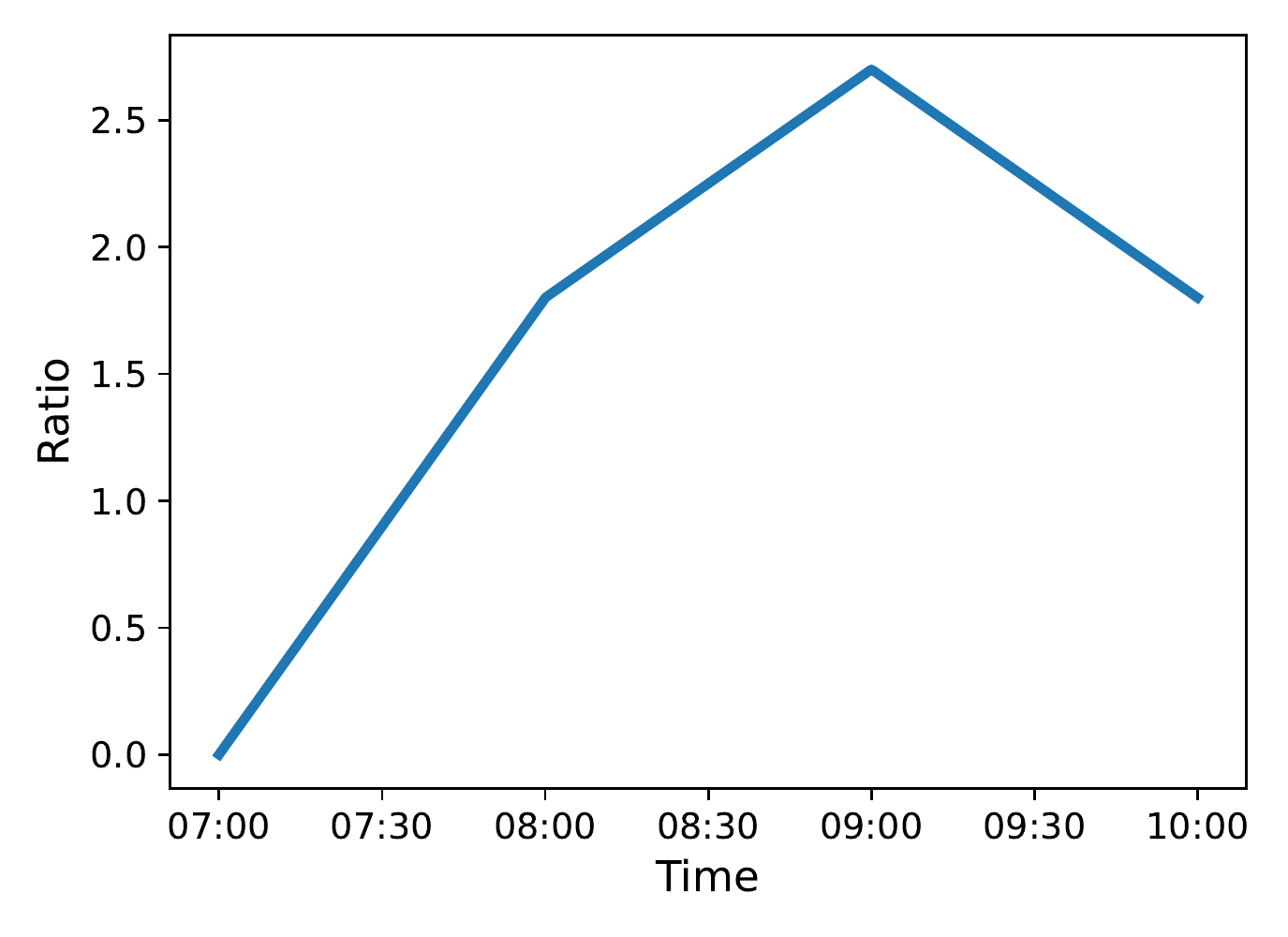}
  }
  \caption{The road and demand information in the case study \#1.}
  \label{fig:network_info01}     
\end{figure}

In this section, a case study on a small network is presented to evaluate the performance of the proposed control method. The structure of the small network is shown in Figure~\ref{fig:study01_network}. There is one 3 km-long freeway with 4 on-ramps and 4 local urban areas in this network. We model the traffic dynamic of urban areas using the bathtub model, and the freeway and on-ramps are modeled using the ACTM. In bathtub models, the space-mean accumulation-speed diagram is modeled with a macroscopic Underwood’s model with a free-flow speed of 90 km/h and a critical accumulation of 1,265 vehicles. The MFD with macroscopic Underwood's model is shown in Figure~\ref{fig:mfd01}. 

For the travel demand, we aim to simulate a 3-hour morning peak from 7:00 AM to 10:00 AM. 
The ratio of the demand pattern is shown in Figure~\ref{fig:demand01}, which means that the travel demand rises from 7:00 AM and reaches the peak at about 9:00 AM, then it falls to the normal demand level when the morning peak ends. The total generated travel demand in the morning peak is 47,271 vehicles.

There are 8 agents in this small network for the 4 on-ramps and 4 perimeters. Due to the high dimension of the joint parameter space of the coordinated ALINEA and Gating, it is unrealistic to simultaneously search the best parameters for each ALINEA and Gating to achieve a global minimal TTT. 
Hence, the optimal demonstrators' parameters of ALINEA and Gating for 4 on-ramps and 4 perimeters are finetuned using the grid search method respectively.
However, in the proposed DRL method, all agents are simultaneously trained to ensure collaborations among agents to achieve a smaller TTT.

In this study, the TTT is used as the objective of the DRL. Other quantities, including average trip speed and delay, are also important to measure the control effects of the control method. The average trip speed is computed as the mean of average speeds for all trips, and the delay is calculated as the ratio of the gaps between free-flow TTT and current TTT over the free-flow TTT. 
Furthermore, we compared the proposed model with the no control, demonstrators (ALINEA or Gating), and other baseline methods, such as DAgger, DQN and DRQN. DAgger \citep{dagger} is an imitation learning algorithm that iteratively learns deterministic policies from a prepared generated dataset. DQN \citep{dqn} is a classic DRL method for discrete control and fully observable problem. DRQN \citep{drqn} is a DRL method focusing on solving discrete control and POMDPs using historical information.

\subsubsection{Experimental results and comparisons}
In addition to the coordinated ramp metering and perimeter control, we also set 2 other control scenarios, ramp-only and perimeter-only. In each control scenario, we trained each model 5 times with different random seeds individually to examine the robustness of the proposed model. The performance of the proposed methods and other baseline methods are shown in Table~\ref{tab:results_study01}. 
If ramps and perimeters are not controlled, the normal average travel time for each traveler is 3,783.93 seconds, and the average delay is 6.69, meaning that the actual travel time is 6.69 times the TTT in free-flow conditions. The average speed for each trip is 30.52 km/h. 
Typically, in the coordinated ramp metering and perimeter control, if they are controlled by the proposed method, the TTT can be decreased by 31.3\%, the delay can be reduced to 4.28, and the average trip speed can increase by 6.7\%. The proposed method reaches the minimal TTT and delay in each control scenario, and achieves the highest average speed except in the perimeter-only scenario. 
One can see the proposed DRL can outperform the teachers ({\em i.e.}, demonstrators) by 46.1\% in TTT. 
Compared to DAgger, DQN, and DRQN, the proposed method can not only learn the pre-determined control strategy, but also explore a better policy based on the knowledge it has learned. This demonstrates the necessity of both demonstration and exploration. 
Importantly, the result in coordinated ramp metering and perimeter control is better than the results in the ramp-only and  perimeter-only with respect to the TTT and delay. It indicates that the simultaneous control of freeways and urban areas can indeed enhance traffic efficiency for the entire network.

\begin{table}[h]
    \begin{tabular}{p{0.10\columnwidth}|p{0.13 \columnwidth}|p{0.17\columnwidth}|p{0.18\columnwidth}|p{0.12\columnwidth}|p{0.15\columnwidth}}
    \hline 
    \textbf{Control} & \textbf{Model} & \textbf{Reward} & \textbf{TTT (s)} & \textbf{Delay} & \textbf{Speed (km/h)} \\
    \hline
    No Control & \diagbox[innerwidth=0.13\columnwidth, dir=SW]{} & \diagbox[innerwidth=0.17\columnwidth, dir=SW]{} & 3783.93 & 6.69 & 30.52 \\
    \hline 
    \multirow{5}*{Ramp} & Demonstrator & \; 824.61 & 2959.31 & 5.02 & 33.55 \\
    & DAgger & \; 925.27 $\pm$ 164.11 & 2858.65 $\pm$ 164.11 & 4.81 $\pm$ 0.33 & 33.81 $\pm$ 0.28 \\
    & DQN & \; 848.20 $\pm$ 88.97 & 2935.71 $\pm$ 88.97 & 4.97 $\pm$ 0.18 & 31.96 $\pm$ 1.34 \\
    & DRQN & \; 895.30 $\pm$ 54.45 & 2888.61 $\pm$ 54.45 & 4.87 $\pm$ 0.11 & 31.93 $\pm$ 0.23 \\
    & \textbf{Proposed} & \; \textbf{959.94} $\pm$ 30.07 & \textbf{2823.97} $\pm$ 30.07 & \textbf{4.74} $\pm$ 0.06 & \textbf{33.99} $\pm$ 0.14 \\
    \hline
    \multirow{5}*{Perimeter} & Demonstrator & \; 566.27 & 3217.64 & 5.54 & \textbf{31.66} \\
    & DAgger & \; 132.35 $\pm$ 156.25 & 2651.57 $\pm$ 156.25 & 6.43 $\pm$ 0.31 & 31.01 $\pm$ 0.40 \\
    & DQN & \;\;\; 89.78 $\pm$ 330.28 & 3694.13 $\pm$ 330.28 & 6.51 $\pm$ 0.67 & 30.39 $\pm$ 0.24 \\
    & DRQN & \; 311.87 $\pm$ 66.86 & 3472.05 $\pm$ 66.86 & 6.06 $\pm$ 0.13 & 30.20 $\pm$ 0.92 \\
    & \textbf{Proposed} & \; \textbf{597.48} $\pm$ 13.74 & \textbf{3186.44} $\pm$ 13.74 & \textbf{5.48} $\pm$ 0.01 & 31.48 $\pm$ 0.28 \\
    \hline
    \multirow{5}*{\shortstack{Ramp \& \\ Perimeter}} & Demonstrator 
    & \;\ 811.28 & 2972.64 & 5.04 & 31.06 \\
    & DAgger & \;\ 805.93 $\pm$ 134.80 & 2977.98 $\pm$ 134.80 & 5.05 $\pm$ 0.27 & 31.86 $\pm$ 1.40 \\ 
    & DQN & \; 424.36 $\pm$ 596.62 & 3359.56 $\pm$ 596.62 & 5.83 $\pm$ 1.21 & 29.65 $\pm$ 2.06 \\
    & DRQN & \; 848.12 $\pm$ 247.15 & 2935.80 $\pm$ 247.15 & 4.97 $\pm$ 0.50 & 31.29 $\pm$ 2.90 \\
    & \textbf{Proposed} & \textbf{1185.08} $\pm$ 98.74 & \textbf{2598}.83 $\pm$ 98.74 & \textbf{4.28} $\pm$ 0.20 & \textbf{32.58} $\pm$ 0.96 \\
    \hline
    \end{tabular}
    \caption{The comparison of the control performance with different methods under different scenarios (mean $\pm$ std for 5 random runs).}
    \label{tab:results_study01}
\end{table}

\begin{figure}[h]
    \centering
    \includegraphics[width=.7\textwidth]{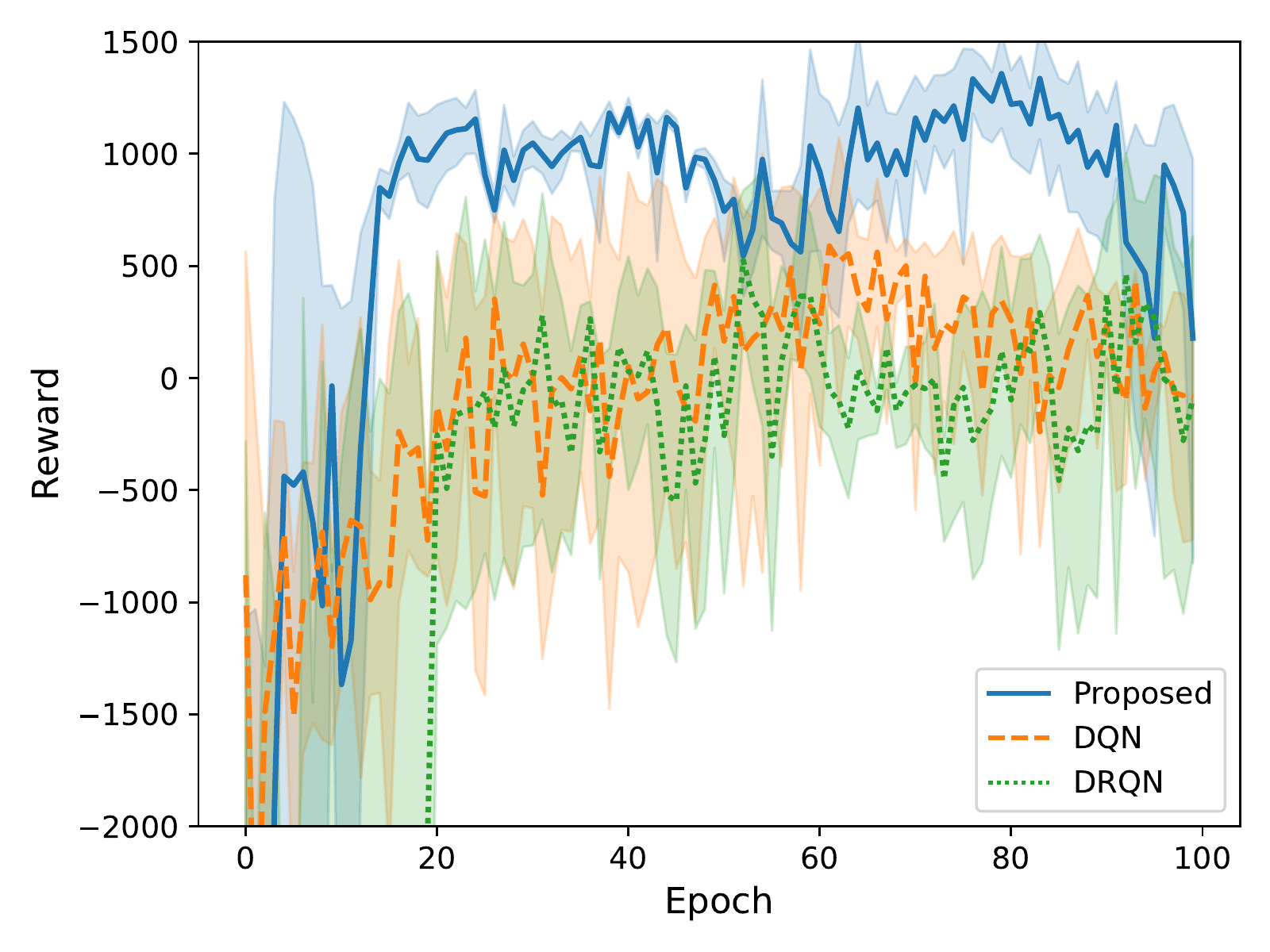}
    \caption{The reward curves of the DQN, DRQN and the proposed method in the training stage.}
    \label{fig:reward_curve}
\end{figure}

The reward curves of the DQN, DRQN, and the proposed DRL method in the training stage are shown in Figure~\ref{fig:reward_curve}. 
It can be seen that the reward of the proposed model is higher than that of DQN and DRQN most time in the training stage. The cumulative reward gradually increased and reached a peak of 1200 at about Epoch \# 80, while the cumulative reward of DQN and DRQN may reach a maximum of 500.

\begin{figure}[h]
    \centering
    \includegraphics[width=1\textwidth]{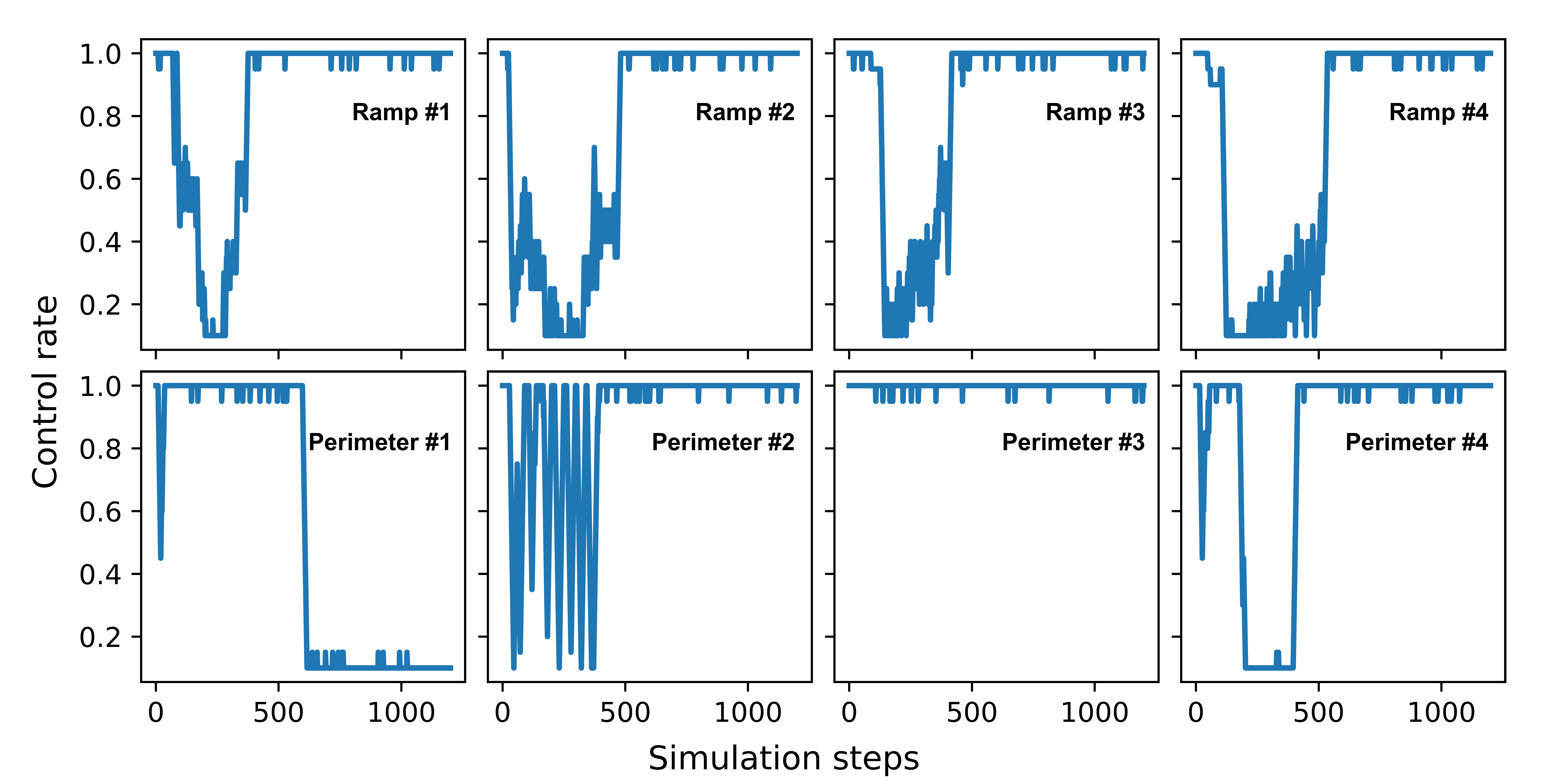}
    \caption{The control rate variation for all controllable ramps and perimeters in case study \#1.}
    \label{fig:action_curve}
\end{figure}

Figure~\ref{fig:action_curve} shows the control rate variation of ramps and perimeters with the proposed method. 
Here we set $\mathbf{u}_{min}$, $\mathbf{U}_{min}$ and $\mathbf{u}_{max}$, $\mathbf{U}_{max}$ as $0.1$ and $1$. It can be seen that all ramps share a similar control pattern where ramps keep open when the freeways are not in congestion, and the metering rate drops drastically to the minimal capacity when ramps are queued to maintain the throughput of freeways. Once the peak hours end, ramps restore to be fully open. The control strategies are different for perimeters. Perimeter \#1 and \#3 remain open, and the control rate of Perimeter \#2 fluctuates when the urban area is in congestion, which resembles the patterns of bang-bang control. It may be interpreted that this urban area is sensitive to accumulation variation. Excess vehicle accumulation may result in serious congestion in downstream freeways and urban areas.

\begin{figure}[h]
  \centering    
  \subfigure[Traffic density on Freeway \#1 with different control methods. (unit: veh/km/lane)] {
  \label{fig:density_freeway01}     
  \includegraphics[width=1\columnwidth]{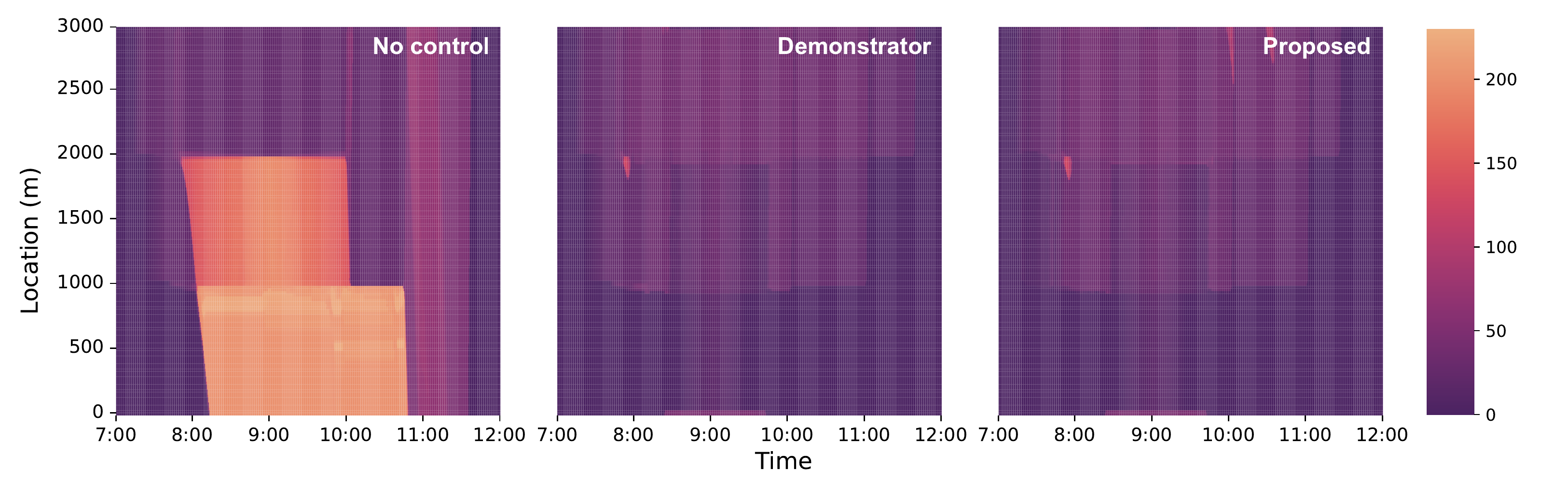} 
  }
  \subfigure[Traffic density on Freeway \#2 with different control methods. (unit: veh/km/lane)] { 
  \label{fig:density_freeway02}     
  \includegraphics[width=1\columnwidth]{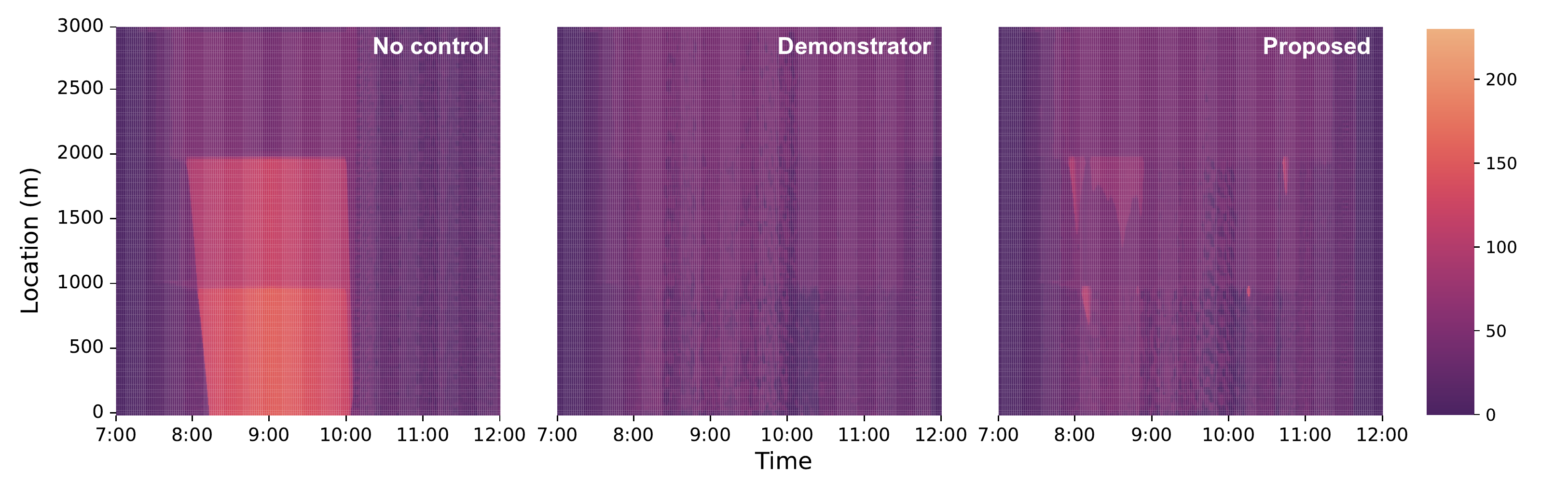}
  }
  \caption{Traffic density on freeways with different control methods. For Freeway \#1 and \#2, there are two ramps located at the 1km and 2km of freeways.}
  \label{fig:density_freeways}     
\end{figure}

\begin{figure}[h]
  \centering    
  \subfigure[Vehicle accumulation in Region \#1] {
   \label{fig:acc_region01}     
  \includegraphics[width=0.45\columnwidth]{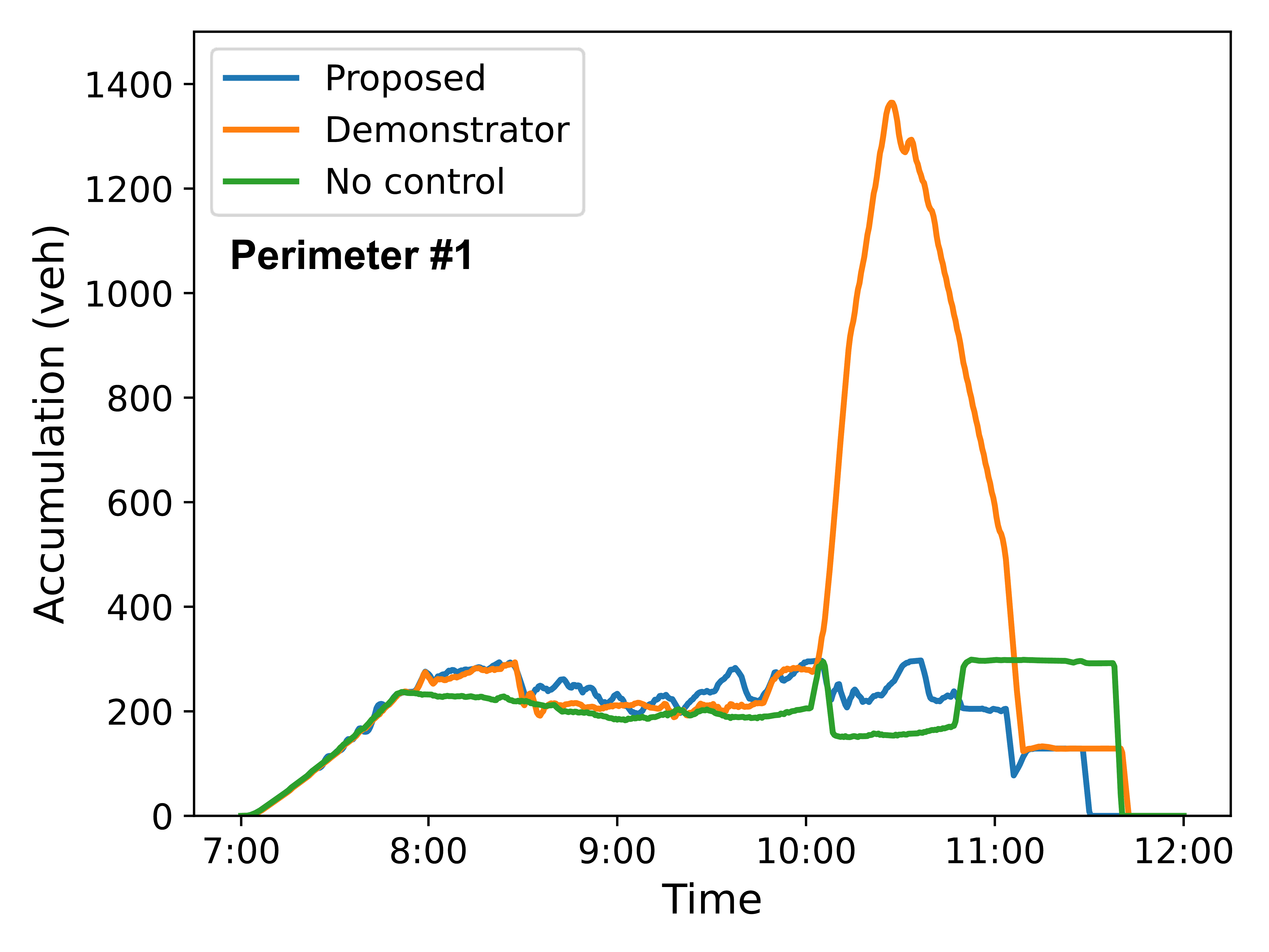}
  }
  \subfigure[Vehicle accumulation in Region \#3] { 
  \label{fig:acc_region03}     
  \includegraphics[width=0.45\columnwidth]{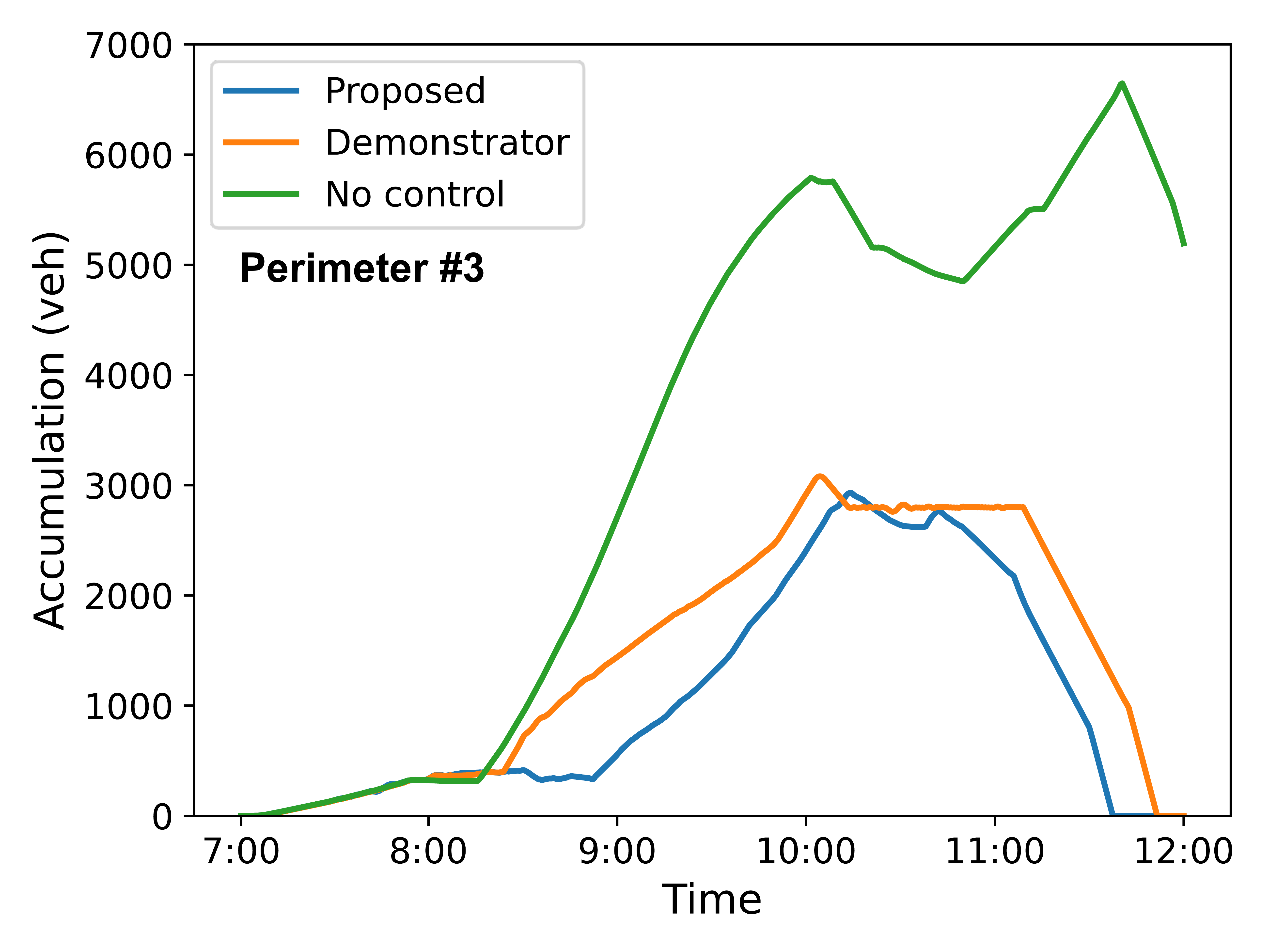}
  }
  \subfigure[Vehicle accumulation in Region \#2] { 
  \label{fig:acc_region02}     
  \includegraphics[width=0.45\columnwidth]{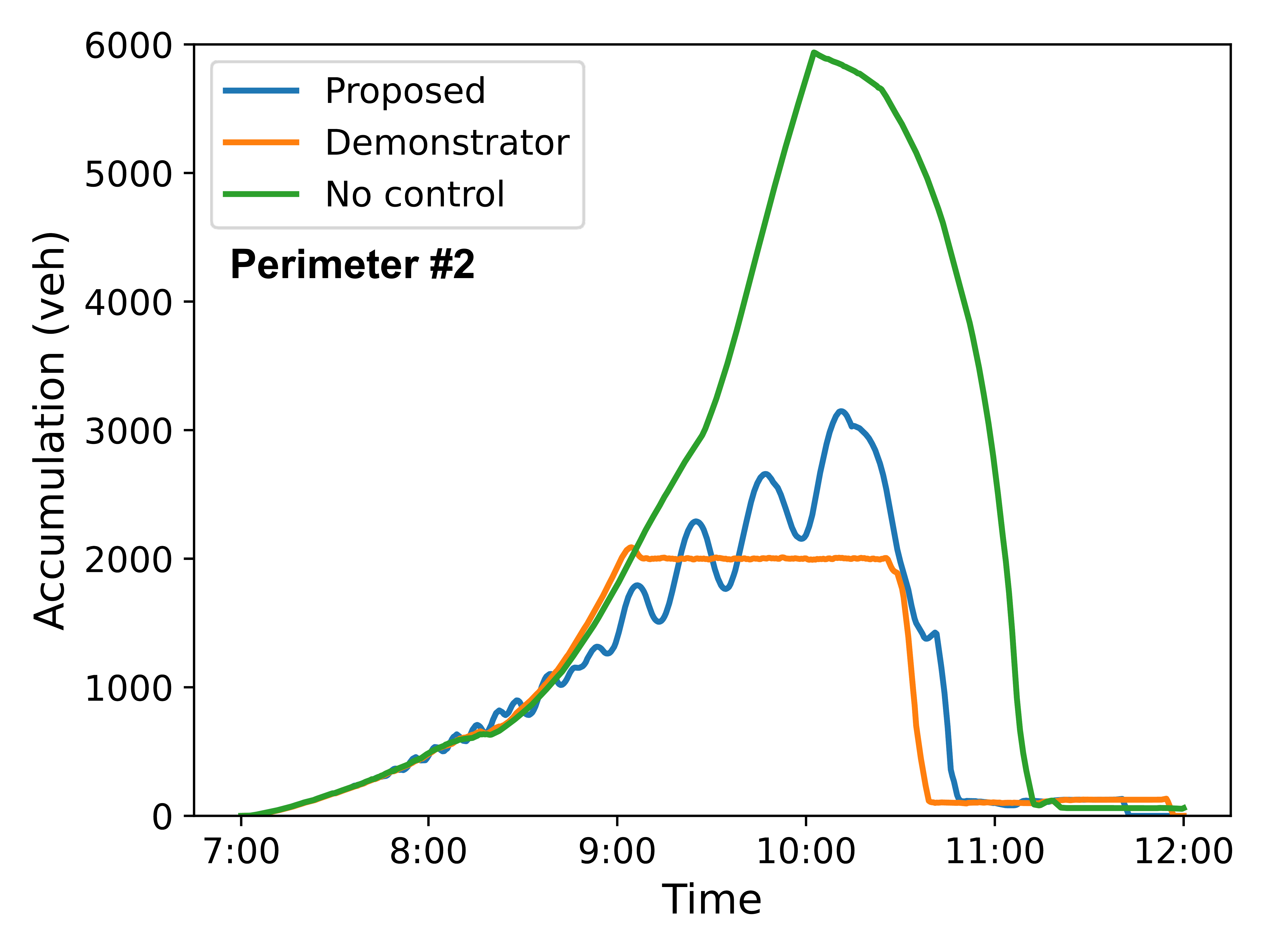}
  }
  \subfigure[Vehicle accumulation in Region \#4] {
  \label{fig:acc_region04}     
  \includegraphics[width=0.45\columnwidth]{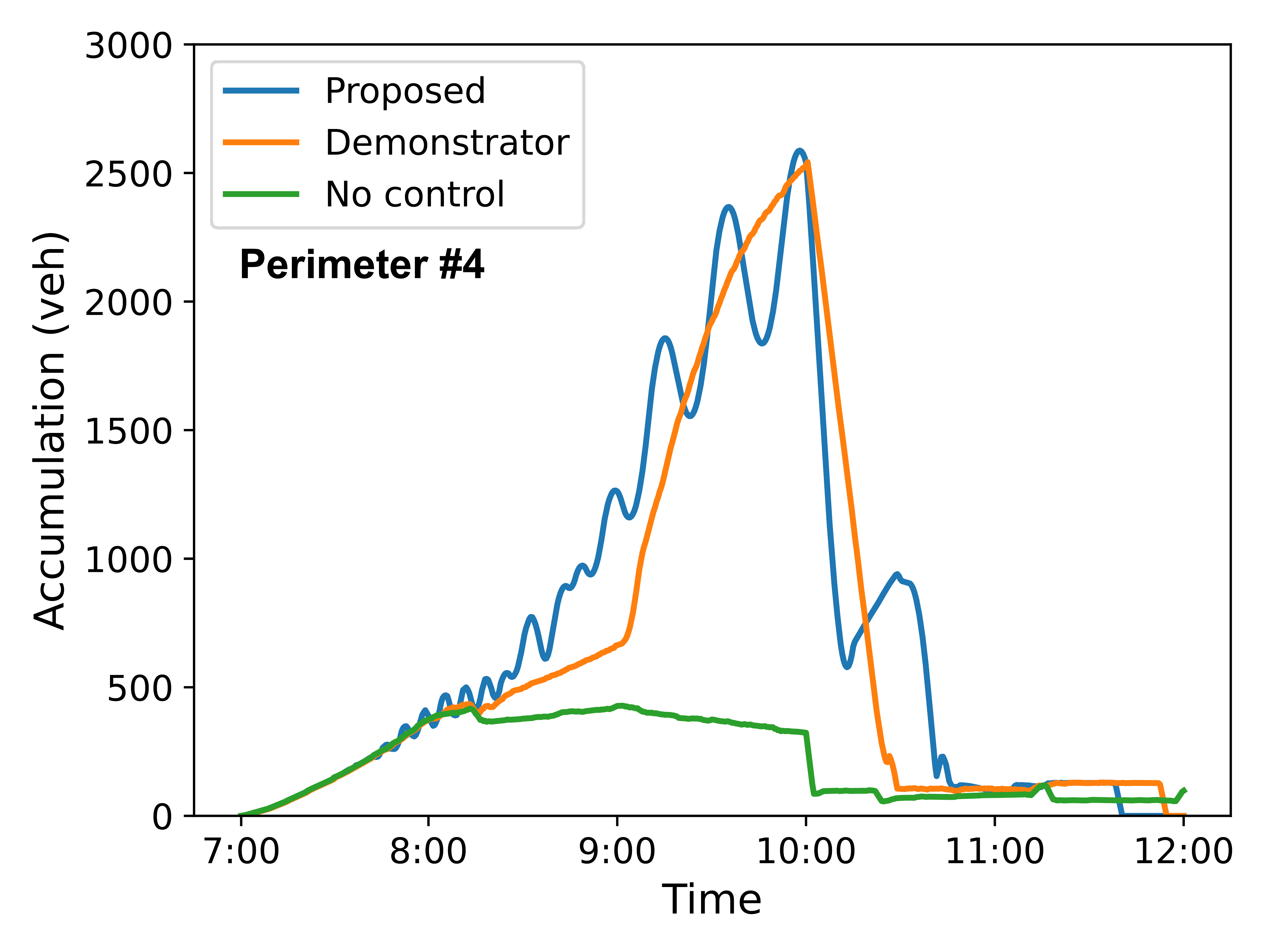}
  }
  \caption{Vehicle accumulation in different regions.}
  \label{fig:acc_regions}     
\end{figure}

The evolution of traffic densities on freeways and vehicle accumulation in urban areas are shown in Figures~\ref{fig:density_freeways} and \ref{fig:acc_regions}. 
In order to better visualize the result, we select 2 road segments on the freeway and named them as Freeway \#1 and Freeway \#2, respectively. Ramp \#1 and \#2 are located at the first 1km and 2km of the Freeway \#1 while Ramp \#3 and \#4 are at the first 1km and 2km of the Freeway \#2. 
Figure~\ref{fig:density_freeway01} shows the evolution of traffic density on Freeway \#1 with no control regime, demonstrator (\textit{i.e.}, ALINEA) and the proposed method. It can be seen that if ramps are not controlled, initially, the congestion emerges at Ramp \#2 from around 8:00 AM, then it spills back rapidly to Ramp \#1. At around 8:20 AM, the congestion spreads to the origin of the freeway. This scenario is highly congested since the traffic density in peak hours surpasses 200 veh/km/lane from the origin of the freeway to Ramp \#1, and surpasses 150 veh/km/lane from Ramp \#1 to \#2. The road segment from Ramp \#2 to the destination of the freeway is not affected by the traffic congestion at ramps since vehicles are most blocked at Ramp \#1 and Ramp \#2, and no congestion spills back from the destination of the freeway. Hence vehicles run with the free-flow speed in this road segment. The congestion from Ramp \#1 to Ramp \#2 ends at around 10:00 AM, and the accumulated vehicles on mainlines begin to move. However, the congestion from the origin of the freeway to Ramp \#2 extends for extra 40 minutes since the remaining queuing vehicles at Ramp \#1 still affect the traffic throughput on the mainline. All vehicles on Freeway \#1 have been moved out at about 11:40 AM, and the congestion time lasted for about 3 hours. 
If Ramp \#1 and \#2 are controlled by the demonstrator or the proposed method, it can be seen that congestion rarely exists on Freeway \#1. Most time, the traffic density will not exceed 100 veh/km/lane everywhere on the freeway. It means that control methods can significantly ease traffic congestion and reduce the TTT for each traveler. If ramps are controlled with demonstrators, all vehicles on freeways are wiped out at about 11:40 AM, while all vehicles on freeways leave at about 11:30 AM. In contrast, if ramps are controlled with the proposed DRL method, the clearance time is 10 minutes 
ahead of the case with demonstrators. The evolution of traffic density on Freeway \#2 shares a similar case with the one on Freeway \#1, and detailed descriptions are omitted.

The vehicle accumulation in all regions is shown in Figure~\ref{fig:acc_regions}. If perimeters are not controlled, vehicles are prone to retain in Region \#2 and Region \#3. In Region \#2, the level of congestion reaches a peak at around 10:00 AM and 6,000 vehicles are trapped in the region. Note that the vehicle accumulation in Region \#2 and Region \#3 is far beyond the total road capacity in a region (3,200 - 4,000 per region), causing a vicious circle of traffic congestion. Based on the MFD, if the current vehicle accumulation exceeds the critical vehicle accumulation, the average speed for all vehicles in the region decreases, and the outflow from the current region reduces consequently. However, the inflow vehicles are not limited unless the vehicle accumulation reaches the boundary capacity. The newborn trips will enlarge the vehicle accumulation and deteriorate the average speed again, which forms a circle. Moreover, the inconsistent uniform distribution of vehicle accumulation among all regions restricts the efficiency of the whole network. While Region~\#2 and Region \#3 are in over-saturation, a few vehicles run in Region \#1 and Region \#4. If parts of vehicles in Region \#2 and \#3 can be balanced in Region \#1 and Region \#4 temporally, the traffic efficiency in the whole network may be enhanced. If perimeters are controlled by the demonstrator or the proposed method, Region \#4 helps to withhold parts of vehicles so that Region \#2 and \#3 can achieve a higher traffic throughput. Controlled by the demonstrator, Region \#1 even helps to retain parts of vehicles. However, the demonstrators did not fulfill a better control performance which may be due to the impact of Freeway \#1 and \#2. The proposed DRL method learns an implicit collaboration between ramps and perimeters which makes the traffic more efficient. 

An interesting observation is that the variation of vehicle accumulation in Region \#2 and Region \#4 under the control of the proposed method forms the shape of waves, which is caused by the variation of control rate in Region \#2. In Figure~\ref{fig:action_curve}, the proposed method outputs a control strategy that resembles bang-bang control in Region \#2. If Region \#2 only allows the minimal traffic flow, Region \#4, as an upstream region, will accumulate vehicles, and Region \#2 will eject vehicles in the opposite manner. The variation of vehicle accumulation forms an ascending wave from 8:00 AM to 10:00 AM due to the newborn trips, and the follow-up phases form the opposite wave. 
The vehicle accumulation in Region \#1 and Region \#3 is not affected by Region \#2 since Region \#2 works like a buffer that flattens the traffic outflow into Region \#1.

\subsubsection{Ablation study}
To validate the function of each component in the proposed DRL method, we present three groups of ablated models: no multi-step returns, no demonstrators, and no multi-step returns with demonstrators. 
The results for each group are shown in Table~\ref{tab:ablation}. Both the demonstrator and multi-step returns are essential to the proposed model. However, the utilization of demonstrators is more important than that of multi-step returns, which further verifies the necessity of demonstrators.
The control performance is further deteriorated if neither demonstrator nor multi-step returns are used, which is equal to the DRQN model.

\begin{table}[h]
    \begin{tabular}{p{0.23\columnwidth}|p{0.16\columnwidth}|p{0.16\columnwidth}|p{0.16\columnwidth}|p{0.16\columnwidth}}
    \hline 
    \textbf{Model} & \textbf{Reward} & \textbf{TTT} & \textbf{Delay} & \textbf{Speed} \\
    \hline 
    Proposed & \textbf{1185.08} $\pm$ 98.74 & \textbf{2598}.83 $\pm$ 98.74 & \textbf{4.28} $\pm$ 0.20 & \textbf{32.58} $\pm$ 0.96 \\
    No nsteps & 1125.91 $\pm$ 105.36 & 2658.00 $\pm$ 105.36 & 4.40 $\pm$ 0.21 & 31.97 $\pm$ 0.49 \\
    No demonstrator & \; 892.21 $\pm$ 171.41 & 2891.70 $\pm$ 171.41 & 4.88 $\pm$ 0.34& 30.01 $\pm$ 1.86 \\
    No demonstrator \& nsteps & \; 848.12 $\pm$ 247.15 & 2935.80 $\pm$ 247.15 & 4.97 $\pm$ 0.50 & 31.29 $\pm$ 2.90 \\
    \hline
    \end{tabular}
    \caption{The ablation study for the coordinated control in the small network (mean $\pm$ std for 5 random runs).}
    \label{tab:ablation}
\end{table}

\subsubsection{Sensitivity analysis}
\begin{figure}[h]
  \centering    
  \subfigure[Total Travel Time] {
   \label{fig:sensitivity_ttt}     
  \includegraphics[width=0.34\columnwidth]{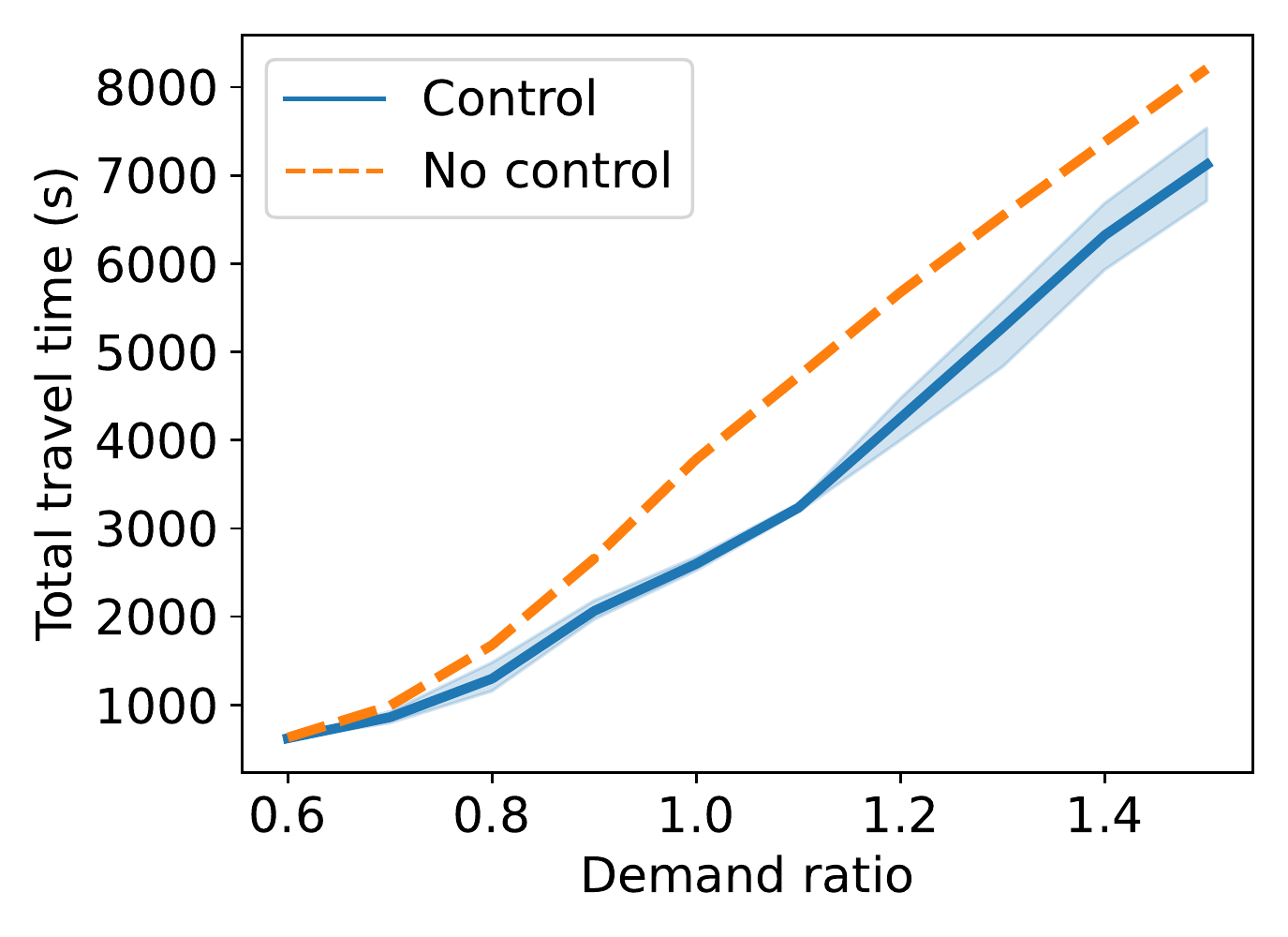} \hspace*{-1.3em}
  }
  \subfigure[Delay] { 
  \label{fig:sensitivity_delay}     
  \includegraphics[width=0.34\columnwidth]{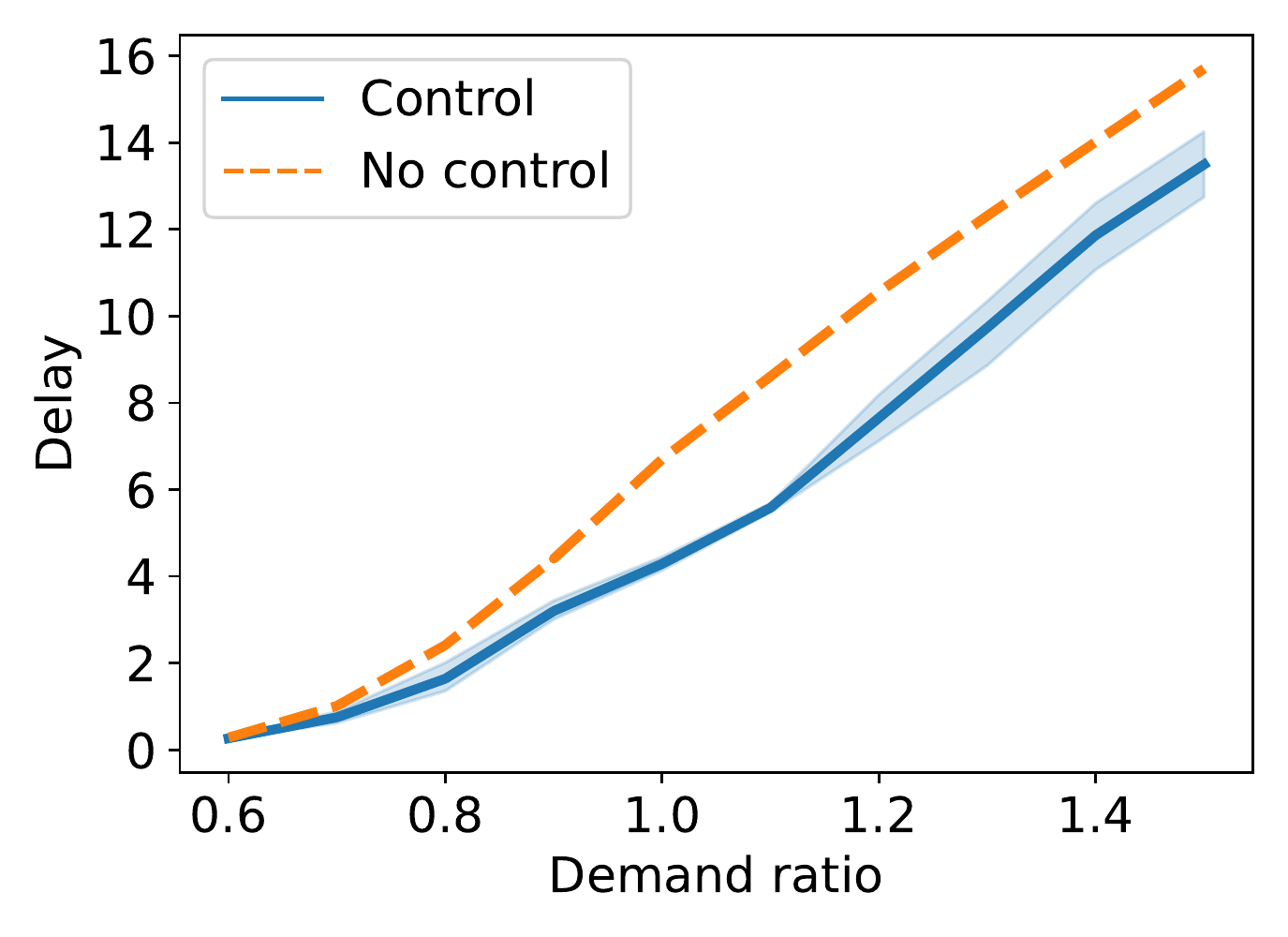} \hspace*{-1.3em}
  }
  \subfigure[Speed] { 
  \label{fig:sensitivity_speed}     
  \includegraphics[width=0.34\columnwidth]{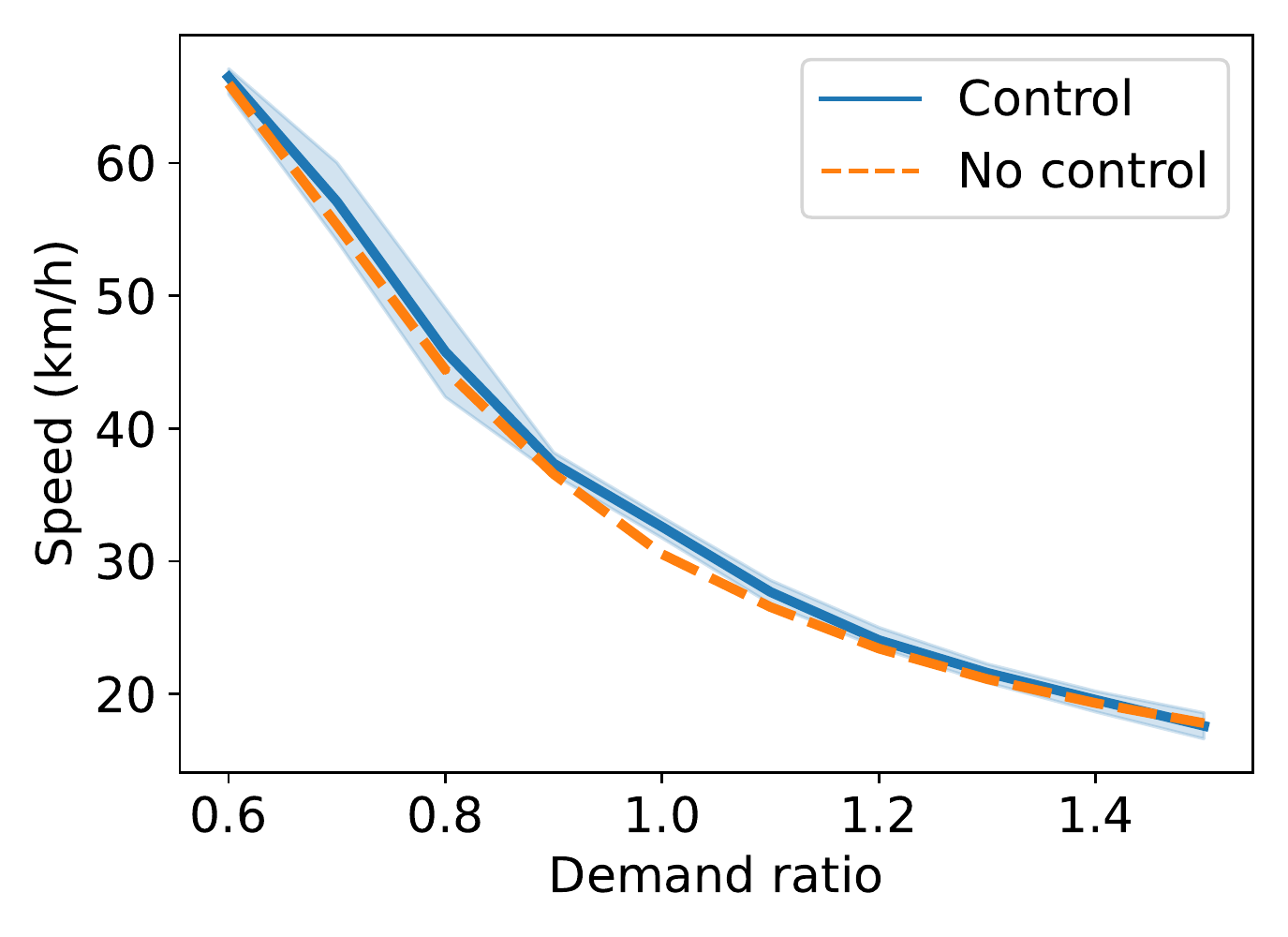}  \hspace*{-1.3em}
  }
  \caption{Sensitivity analysis with different demand profiles.}
  \label{fig:sensitivity}     
\end{figure}

When we enact a control strategy in an urban network, the robustness of the control method is one of the most important issues to consider since the daily travel demand may vary randomly and the proposed method should be suitable for various traffic scenes. Hence we test the proposed model trained with the normal demand profile directly in different demand profiles without extra training 5 times. In different demand profiles, the OD pairs keep unchanged, while the volume in each demand record ranges from 60\% to 150\% of the normal demand. The comparison of performances between the control and no control methods are shown in Figure~\ref{fig:sensitivity}. When the travel demand reduces to 60\% of the normal demand in the morning peak, it can be seen that there is nearly no difference in TTT, delay and speed between the control and no control methods. 
However, when the travel demand increases, the gaps in TTT and delay between the no control and the proposed method enlarges gradually and reaches the maximum when the demand ratio is 1.1. Then two lines of TTT and delay remain parallel from 1.1 to 1.5 since the traffic is over-congested and the traffic control efficiency is stablized. 

\subsection{Case study \#2: A real-world large-scale network}
\label{sec:case_study02}
To further evaluate the performance of the proposed DRL method on a large-scale network, we implement the coordinated ramp metering and perimeter control on a real-world road network in the Kowloon district, Hong Kong SAR, China. The non-stationary properties of the environment in large-scale networks may be more challenging for the DRL method than in small networks. Hence the training epoch is set to 200, and other hyperparameters are the same as those in case study \#1.
The preparation for different parameter settings for the real-world network model is introduced as follows.
\begin{figure}[h]
  \centering    
  \subfigure[The Kowloon District in the OpenStreetMap.] {
   \label{fig:kowloon_osm}     
  \includegraphics[width=0.48\columnwidth]{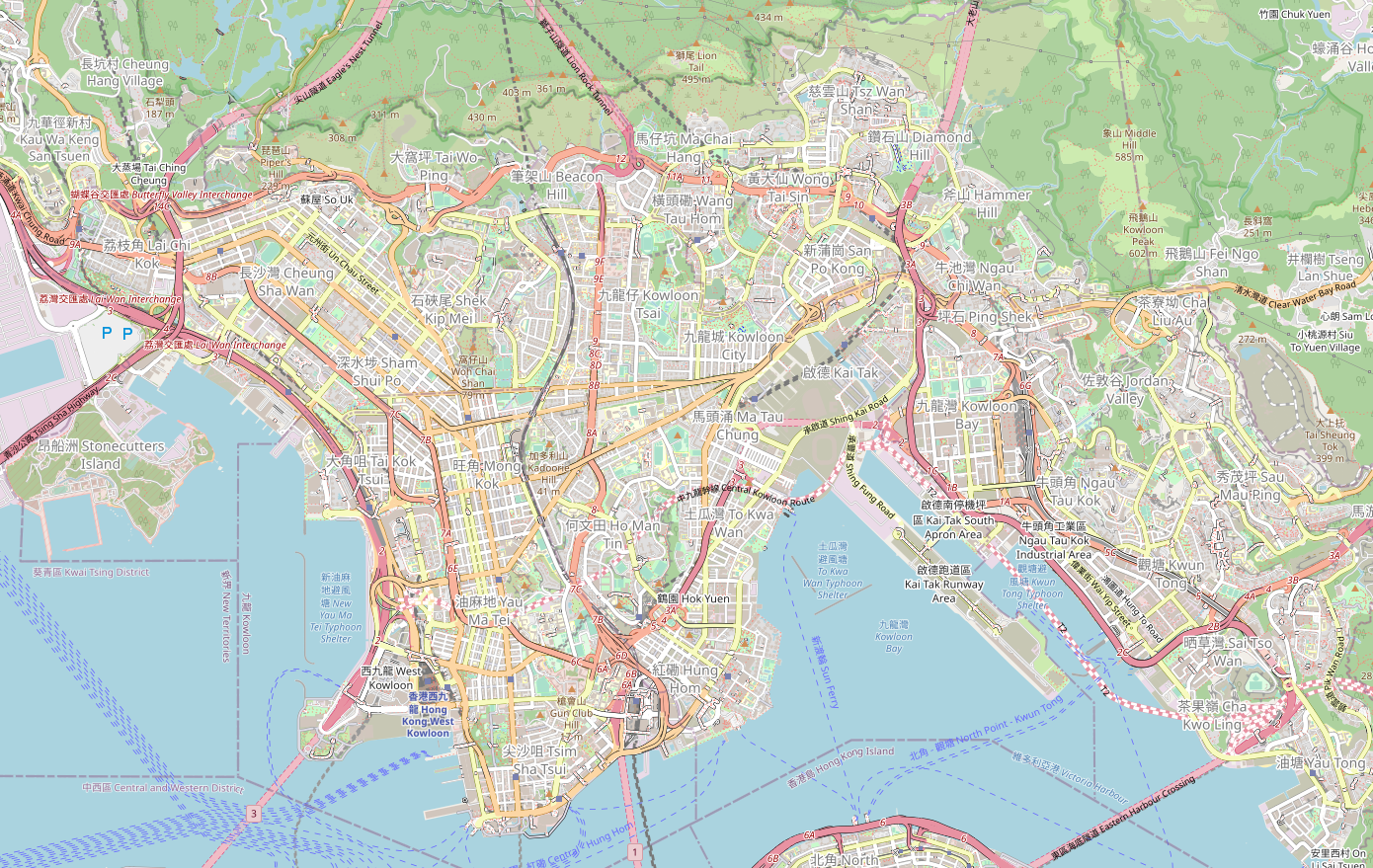}
  }
  \subfigure[The controllable agent in Kowloon District. The blue polygons represent perimeter agents, while red dots mean ramp agents.] { 
  \label{fig:kowloon_agent}     
  \includegraphics[width=0.48\columnwidth]{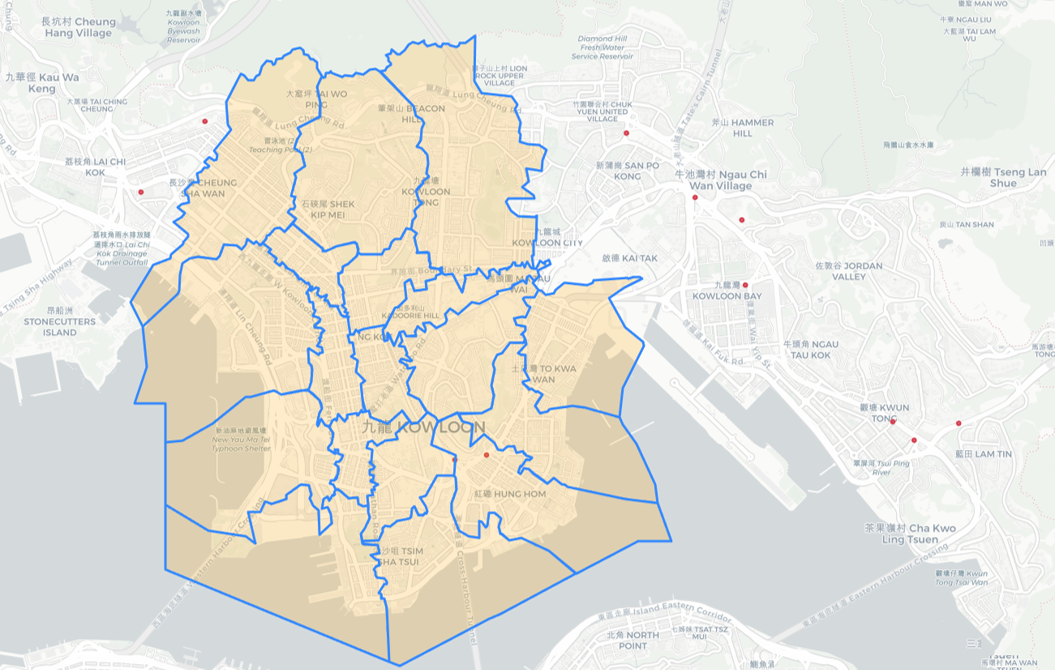}
  }
  \subfigure[The partitioned urban areas simulated with the bathtub model in Kowloon District.] { 
  \label{fig:kowloon_perimeter}     
  \includegraphics[width=0.48\columnwidth]{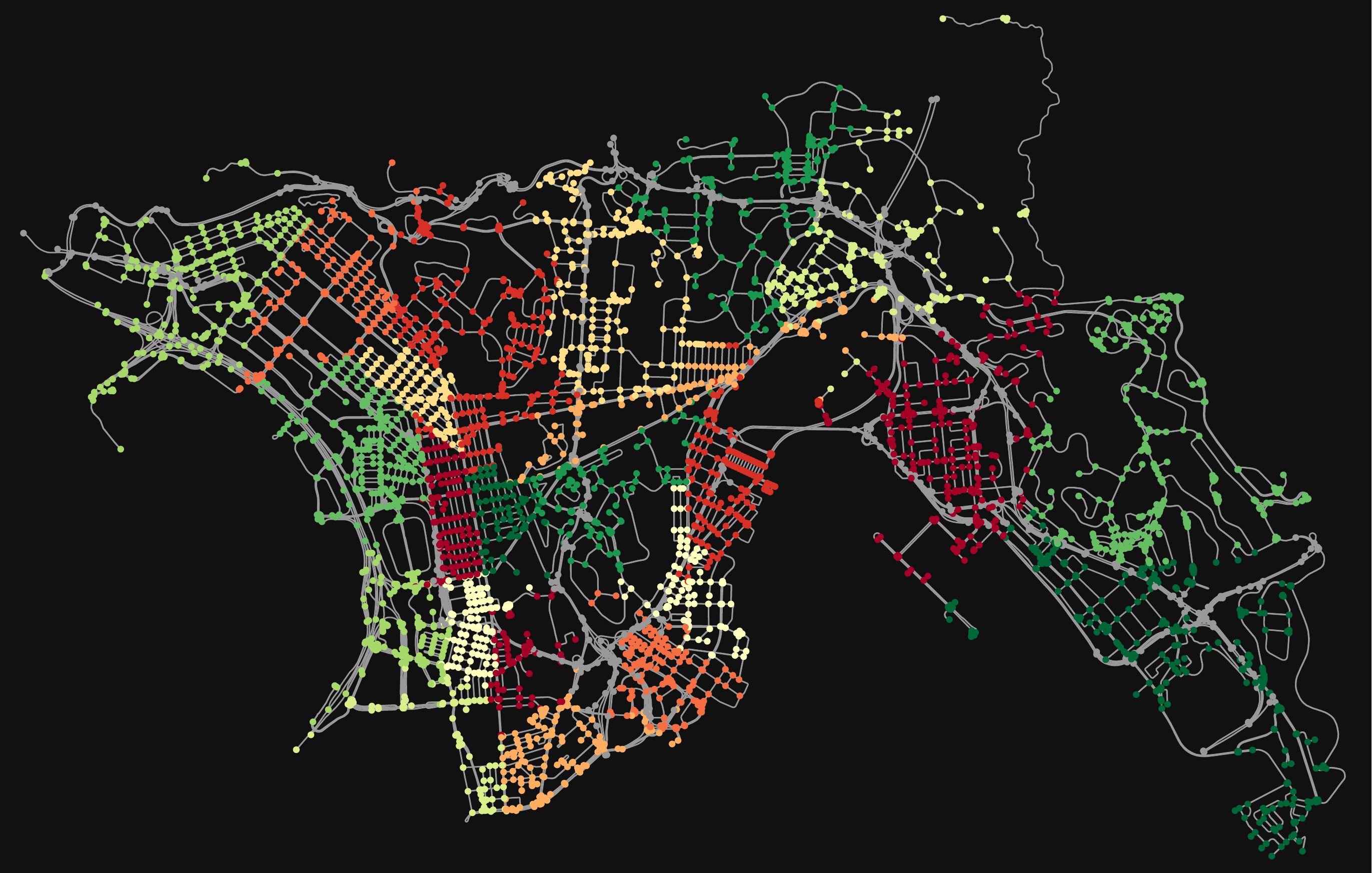}
  }
  \subfigure[The freeways simulated with the ACTM in Kowloon.] { 
  \label{fig:kowloon_highway}     
  \includegraphics[width=0.48\columnwidth]{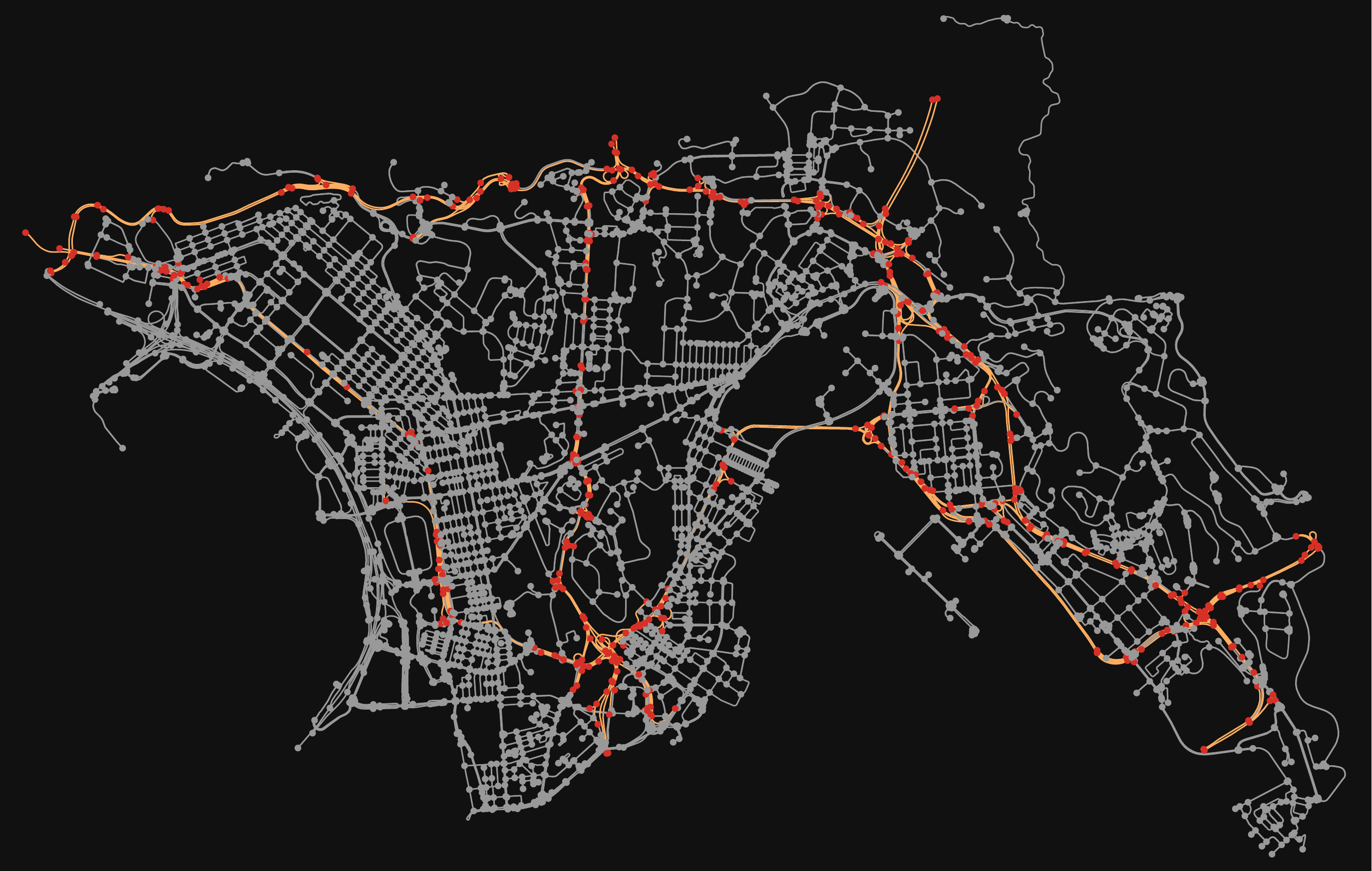}
  }
  \caption{Overview of the study area.}
  \label{fig:flow_regulator}     
\end{figure}

\paragraph{\textbf{Network data}}
The road network is constructed from the OpenStreetMap (OSM, shown in Figure~\ref{fig:kowloon_osm}) with OSMnx \citep{OSMnx}, which can convert the OSM into a directed graph. It can also correct broken links and delete non-junction nodes for network consolidation. The processed road network consists of 4,054 nodes and 7,385 edges. The speed limit and road capacity are acquired based on the road level, lane number, and jam density from the OSM attributes.

\paragraph{\textbf{Network cluster profile}}
The network is partitioned into freeways (including ramps) and urban areas based on the road attributes in the OSM. For urban areas, we further split into several homogeneous regions to ensure a uniform relationship of MFD. The Leiden algorithm \citep{Leiden}, which is a community detection algorithm, is adopted for the split. A total of 29 regions are obtained, in which 28 regions are urban areas simulated with the generalized bathtub model (shown in Figure~\ref{fig:kowloon_perimeter}), and one region contains freeways and ramps that are modeled by the ACTM (shown in Figure~\ref{fig:kowloon_highway}).

\paragraph{\textbf{MFD parameters}}
The MFD for each region is acquired through the regional network topology. Recent work \citep{MFD_calibration} reveals a relationship between the topology of the network and the macroscopic Underwood's model using the taxi data in the Kowloon district, which is formulated in Equation~\ref{eq:mfd_topo}.
\begin{equation}
  v = 53.874 \cdot \exp(-0.077 \gamma_d) \cdot \exp \left( \frac{-k}{3.161 \times 10^6 / Deg_t} \right),
  \label{eq:mfd_topo}
\end{equation}
\noindent  where $v$ is the average density and $k$ is the vehicle density in this region,	$\gamma_d$ represents the number of junctions per kilometer, and degree density $Deg_t$ represents the average node degree per square kilometer.

\begin{figure}[h]
    \centering
    \includegraphics[width=1\textwidth]{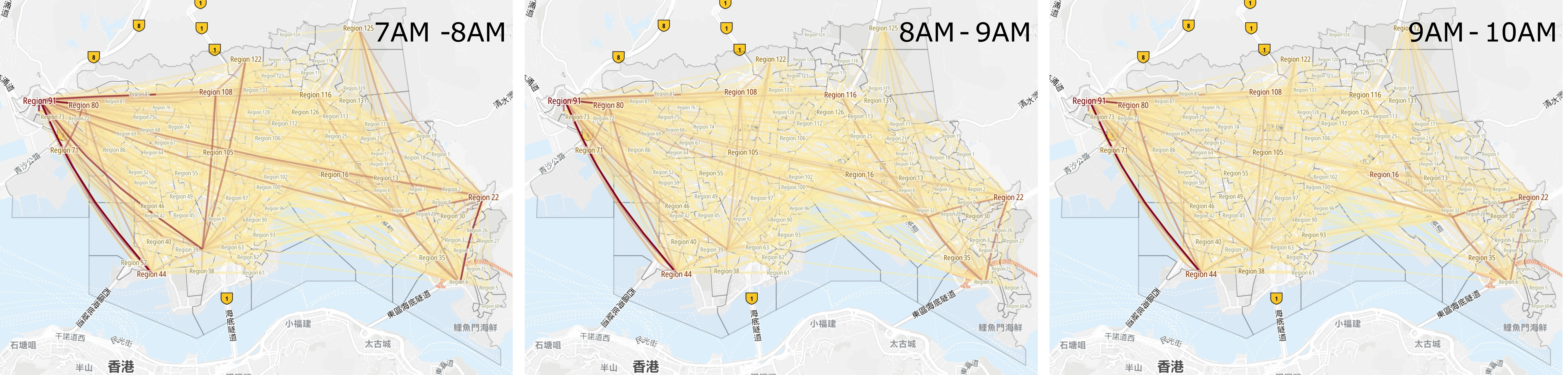}
    \caption{The hourly OD demand profile from 7:00 AM to 10:00 AM. Darker color indicates more trips between the OD pair.}
    \label{fig:od_demand}
\end{figure}

\paragraph{\textbf{Time-of-day OD demand}}
The time-of-day OD demand data is requested from the TomTom Move\footnote{\url{https://move.tomtom.com}}. The study area contains 135 OD regions, and we use the average OD demand in the morning peak (7:00 AM to 10:00 AM) for 30 days as the demand. There are a total of 228,306 trips in the morning peaks, and the visualization of the spatial OD flow for each hour is shown in Figure~\ref{fig:od_demand}. 


\subsubsection{Experimental results}
There are 29 agents in the study area: 11 of them are ramps and 18 of them are perimeters. The location of ramps and perimeters are shown in Figure~\ref{fig:kowloon_agent}. The blue borough represents the perimeter and the red dot denotes the ramp. The comparison of the control performance with different methods is shown in Table~\ref{tab:results_study02}. The average travel time is around 33 minutes without any control method. If ramps and perimeters are controlled with the proposed method, each traveler can save extra 3 minutes on roads, accounting for a 7.8\% reduction in TTT and a 6.1\% increase in average trip speed. Compared to demonstrators (ALINEA and Gating), the proposed DRL method can reduce the TTT further by 12.9\%. On the scope of the real-world and large-scale network, the proposed control method significantly improves the network efficiency.
\begin{table}[h]
    \begin{tabular}{p{0.10\columnwidth}|p{0.13\columnwidth}|p{0.17\columnwidth}|p{0.18\columnwidth}|p{0.12\columnwidth}|p{0.15\columnwidth}}
    \hline 
    \textbf{Control} & \textbf{Model} & \textbf{Reward} & \textbf{TTT} & \textbf{Delay} & \textbf{Speed} \\
    \hline
    No Control & None &\diagbox[innerwidth=0.17\columnwidth, dir=SW]{} & 2059.38 & 10.15 & 23.10 \\
    \hline 
    \multirow{5}*{\shortstack{Ramp \& \\ Perimeter}} & Demonstrator
    & 142.40 & 1896.21& 9.27 & 24.50 \\
    & DAgger & 135.75 $\pm$7.70 & 1903.79 $\pm$ 8.78 & 1.21 $\pm$ 0.01 & 24.13 $\pm$ 0.35 \\ 
    & DQN & 132.46 $\pm$9.82 & 1907.54 $\pm$ 11.19 & 1.22 $\pm$ 0.01 & 23.21 $\pm$ 0.32 \\
    & DRQN & 140.53 $\pm$ 8.29 & 1898.48 $\pm$ 9.32 & 1.20 $\pm$ 0.01 & 23.54 $\pm$ 0.47 \\
    & \textbf{Proposed} & \textbf{160.84} $\pm$ 4.00 & \textbf{1875.48} $\pm$ 4.61 & \textbf{1.18} $\pm$ 0.00 & \textbf{24.53} $\pm$ 0.10 \\
    \hline
    \end{tabular}
    \caption{The comparison of the control performance with different methods under different scenarios (mean $\pm$ std for 5 random runs).}
    \label{tab:results_study02}
\end{table}


\clearpage
\section{Conclusion}
\label{sec:conclusion}
In this paper, we propose a meso-macro dynamic traffic network model and a demonstrated-guided DRL method for coordinated ramp metering and perimeter control in large-scale networks. 
For the meso-macro traffic modeling, we integrate the ACTM and the generalized bathtub model to achieve multi-resolution and high efficiency  in modeling large-scale networks. 
For the coordinated control, when we aim to control multiple agents in a complex environment simultaneously, the non-stationary environment makes it difficult for the DRL method to learn an effective policy. Hence we leverage demonstrators, (\textit{e.g.}, concise traditional controllers) to guide the DRL methods to stabilize the training of multiple agents.

The proposed DRL-based control framework has been evaluated in two case studies, a small network and a real-world and large-scale network in the Kowloon district, Hong Kong. For both case studies, the proposed method is compared with existing baselines, and it outperforms the demonstrators and existing DRL methods. In case study \#1, the result shows that it can save 31.3\% TTT for each traveler. We further conduct an ablation study and a sensitivity analysis to validate the function of each component and the robustness of the proposed DRL method. In case study \#2, we apply the proposed method in a real-world large-scale network. The results show that it can save 7.8\% TTT for each traveler, and compared to the demonstrator, the control performance can be further enhanced by 12.9\% based on the TTT. Both case studies demonstrate the great potential of using demonstrators' guidance in DRL training.

In future research, more extensions regarding the DRL method can be made based on the current study. For example, we can consider the explicit communication between ramps and perimeters in a large-scale control, which is important in Multi-agent Reinforcement Learning (MARL), and each agent may further explore a better control strategy by receiving messages from other agents. 
A simple DRL method can be the demonstrator of complex DRL methods, so we can adopt the concept of knowledge distillation to design a self-evolving DRL-based controller.
Moreover, it would be practical to study whether the developed DRL can be generalizable for different network topologies and traffic scenarios in the real world.

\section*{Acknowledgment}
The work described in this paper was supported by grants from the Research Grants Council of the Hong Kong Special Administrative Region, China (Project No. PolyU/25209221 and PolyU/15206322), a grant from the Research Institute for Sustainable Urban Development (RISUD) at the Hong Kong Polytechnic University (Project No. P0038288), and a grant from the Otto Poon Charitable Foundation Smart Cities Research Institute (SCRI) at the Hong Kong Polytechnic University (Project No. P0043552).

\clearpage
\bibliography{citation}
\clearpage
\appendix
\section{Notations}
\label{apx:notations}
\begin{longtable}{p{0.15\columnwidth}|p{0.80\columnwidth}}
\hline
\multicolumn{2}{l}{\textbf{Meso-macro traffic modeling}} \\
\hline
$\mathbf{G}$ & The traffic network. \\ 
$\mathbf{G}_i$ & The network of region $i$. \\ 
$\mathbf{V}$ & The node set of the traffic network. \\ 
$\mathbf{V}_i$ & The node set of region $i$ in the traffic network. \\ 
$\mathbf{E}$ & The link set (\textit{road set}) of the traffic network. \\ 
$\mathcal{N}_d$ & The number of regions in the traffic network. \\
$e_{r,s}$ & The road from node $r$ to node $s$. \\
$l_{r,s}$ & The length of road $l_{r,s}$. \\
$\mathcal{D}_{r,s}(t)$ & The demand of road $e_{r,s}$. \\
$\mathcal{S}_{r,s}(t)$ & The supply of road $e_{r,s}$. \\
$\delta_{r,s}$ & The length of cell on road $e_{r,s}$. \\
$v_{r,s}^{max}$ & The free-flow speed of road $e_{r,s}$. \\
$n_{r,s}^{k}(t)$ & The vehicle number in the $k$th cell of road $e_{r,s}$. \\
$\hat{n}_{r,s}(t)$ & The jam density of road $e_{r,s}$. \\
$f_{r,s}^{k,k+1}(t)$ & The internal flow from cell $k$ to cell $k+1$ on road $e_{r,s}$. \\
$R_{r,s}^{k}(t)$ & The on-ramp flow on the $k$th cell of road $e_{r,s}$. \\
$S_{r,s}^{k}(t)$ & The off-ramp flow on the $k$th cell of road $e_{r,s}$. \\
$\mu_{r,s}(t)$ & The number of trip begun on road $e_{r,s}$. \\
$\nu_{r,s}(t)$ & The number of trip ended on road $e_{r,s}$. \\
$\phi_{*,r,s}(t)$ & The external traffic flow into road $e_{r,s}$. \\
$\phi_{r,s,*}(t)$ & The exteral traffic flow from road $e_{r,s}$. \\
$\phi(\cdots)$ & The external flow allocation that regulates traffic flow between different roads. \\
$\psi_{+}(r)$ & The downstream nodes of node $r$. \\
$\psi_{-}(r)$ & The upstream nodes of node $r$. \\
$w_{r,s}$ & The spillback speed on road $e_{r,s}$. \\
$q_{r,s}^{max}$ & The capacity of road $e_{r,s}$. \\
$\gamma$ & The on-ramp flow blending parameter. \\
$\zeta$ & The on-ramp flow allocation coefficient. \\
$c_{r',s'}(t)$ & The capacity for on-ramp flow. \\
$\rho_{r,s}^{k}$ & The ramp rate of the $k$th cell on road $e_{r,s}$. \\
$\beta_{r,s}^{k}(t)$ & The off-ramp split factor of the $k$ cell on road $e_{r,s}$. \\
$\overline{N}_d(t)$ & The vehicle accumulation in region $d$. \\
$\overline{\Mu}_d(t)$ & The trip begun in region $d$. \\
$\overline{\Nu}_d(t)$ & The trip finished in region $d$. \\
$\overline{\Phi}_{d,*}(t)$ & The external traffic flow from region $d$. \\
$\overline{\Phi}_{*,d}(t)$ & The external traffic flow to region $d$. \\
$\Mu_d(t, \xi)$ & The trip begun in region $d$ with remaining distance of $\xi$. \\
$\Phi_{*,d}(t, \xi)$ & The external traffic flow from region $d$ with remaining distance of $\xi$.\\
$V_{d}(t)$ & The average speed in region $d$.\\
$\hat{V}_{d}(\cdot)$ & The MFD of region $d$. \\
$L_{d}^{sum}$ & The total road length in region $d$. \\
$L_{d}^{max}$ & The length of the longest trajectory in region $d$. \\
$\hat{\Phi}(\cdots)$ & The external traffic allocator that regulates the traffic flow between different regions. \\
$\mathfrak{D}_{d}(t)$ & The demand of region $d$. \\
$\mathfrak{S}_{d}(t)$ & The supply of region $d$. \\
$\Psi_{+}(d)$ & The downstream regions of region $d$. \\
$\Psi_{-}(d)$ & The upstream regions of region $d$. \\
$\hat{N}_d$ & The jam vehicle accumulation in region $d$. \\
$C_{d,+}^{margin}$ & The outflow capacity at the boundary of region $d$. \\
$C_{d,-}^{margin}$ & The inflow capacity at the boudnary of region $d$. \\
$\Rho_d(t)$ & The control rate of the perimeter of region $d$. \\
\hline
\multicolumn{2}{l}{\textbf{DRL method}} \\
\hline
$\mathcal{X}$ & The state set. \\
$\mathcal{U}$ & The action set.\\
$\mathcal{P}$ & The state transition probability. \\
$\mathcal{R}$ & The reward set. \\
$\Omega$ & The observation set for an agent. \\
$\mathcal{O}$ & The observation probability in the environment. \\
$\mathbf{x}_t$ & The state. \\
$\mathbf{o}_t$ & The observation. \\
$\mathbf{h}_t$ & The historical information. \\
$\mathbf{u}_t$ & The action.  \\
$\mathbf{r}_t$ & The reward. \\
$\hat{\mathbf{r}}_t^i(\eta)$ & The multi-steps reward with  a step size of $\eta$. \\
$G_t$ & The discounted cumulative reward from time $t$. \\
$\lambda$ & The discount factor. \\
$\mathcal{N}_p$ & The number of total trips. \\
$T_i^{end}$ & The end time for trip $i$. \\
$T_i^{start}$ & The start time for trip $i$. \\
$T^{term}$ & The termination time for the simulation. \\
$\mathcal{N}^{inj}(t)$ & The number of injected vehicle. \\
$\mathcal{N}^{run}(t)$ & The number of running vehicle.\\
$\mathcal{N}^{com}(t, t')$ & The number of completed vehicle from time $t$ to time $t'$.\\
$C_r$ & A normalization term for reward. \\
$KI_{r,s}^{k}$ & The integral parameter in ALINEA. \\
$\dot{n}_{r,s}^{k}$ & The threshold for the $k$th cell on road $e_{r,s}$ in ALINEA. \\
$KP_{d}$ & The proportional parameter in Gating. \\
$KI_{d}$ & The integral parameter in Gating. \\
$\dot{N}_{d}$ &  The threshold for the region $d$ in Gating. \\
$\eta$ & The steps of rewards. \\
$Q(\cdots)$ & The action value function. \\
$\pi(\cdots)$ & The agent policy. \\
$\overline{\pi}(\cdots)$ & The demonstrator policy. \\
$\pi(\cdots)$ & The dummy policy. \\
$\mathcal{N}_B$ & The replay buffer size. \\
$\mathcal{N}_m$ & The epoch number. \\
$\mathcal{N}_o$ & The period for training DRL methods. \\
$\mathcal{N}_c$ & The period for cloning the source network. \\
$\epsilon_{start}$ & The initial value of the $\epsilon$. \\
$\epsilon_{end}$ & The final value of the $\epsilon$. \\
$\epsilon_{last}$ & The duration of $\epsilon$ decayed in the training process. \\
$\alpha_{min}$ & The minimal value that used to normalize the value of $\alpha_t$. \\
$\alpha_{step}$ & The normalization term for the time $t$ in calculation of $\alpha_t$. \\
$\alpha_{term}$ & The termination steps in calculation of $\alpha_t$. \\
$\alpha_{amp}$ & The amplification of the KL divergence. \\
\hline
\end{longtable}

\section{Conservation law of different cells in ACTM}
\label{apx:conservation_law}
In this section, we present the conservation law of traffic dynamic in source cells, sink cells, general cells and off-ramp cells in the ACTM. To simplify the evolution of traffic flows, we made the following assumptions.
\begin{assumption}
In the ACTM, a trip begins at the source cell and ends at the sink cell.
\end{assumption}

\begin{assumption}
Any source cells and sink cells are not connected with on-ramp entrances or off-ramp exits.
\end{assumption}

Based on these assumptions, the conservation law of traffic flow in source cells, sink cells, general cells, and off-ramp cells can be discussed separately.

\paragraph{The conservation law of a source cell}
In a source cell, there is no internal flow into the source cell since it is the first cell on the road. The conservation law of vehicles in the source cell is formulated in Equation~\ref{eq:con_source},
\begin{equation}
    \begin{aligned}
        n_{r,s}^{1}(t) &= n_{r,s}^{1}(t - 1) + \mu_{r,s}(t) + \phi_{*,r,s}(t) - f_{r,s}^{1,2}(t) \\
        \phi_{*,r,s}(t) &= \sum_{i \in \psi^{-}(r)}\phi_{i,r,s}(t) \\
        \phi_{i,r,s}(t) &= \hat{\phi} \left( \left\{ \mathcal{D}_{j,r}(t) \vert \forall j \in \psi^{-}(r) \right\}, \left\{ \mathcal{S}_{r,k}(t) \vert \forall k \in \psi^{+}(r) \right\} \right), \\
        & \qquad \qquad \qquad \forall r,s \in \mathbf{V}
    \end{aligned}
    \label{eq:con_source}
\end{equation}
\noindent where $n_{r,s}^{1}(t)$ is the vehicle number in the source cell at time $t$, $f_{r,s}^{1,2}$ is the internal traffic flow from the first cell (source cell) to the second cell on road $e_{r,s}$ at time $t$. $\phi_{i,r,s}(t)$ denotes the external traffic flow from road $e_{i,r}$ into road $e_{r,s}$. $\psi_{r}^{-}$ and $\psi_{r}^{+}$ mean sets of upstream and downstream nodes that directly connects to node $r$ respectively. $\hat{\phi}(\cdots)$ represents a function that allocates the external flows between different roads by randomly transferring vehicles from the sink cell of the upstream road to the source cell of the downstream road. $\mathcal{D}_{j,r}$ is demand on road $e_{j,r}$ at time $t$, and $\mathcal{S}_{r,k}$ is the supply on road $e_{r,k}$ at time $t$. The demand and supply are calculated as 
\begin{equation}
    \begin{aligned}
    \mathcal{D}_{r,s}(t) &= \min \left\{ n_{r,s}^{-1}(t) - \nu_{r,s}(t), q_{r,s}^{max} \right\} \\
    \mathcal{S}_{r,s}(t) &= \min \left\{ \frac{w_{r,s}}{v_{r,s}^{max}} \left( \hat{n}_{r,s} - n_{r,s}^{-1}(t) - \mu_{r,s}(t) \right), q_{r,s}^{max} \right\} \\
    \end{aligned}
    \label{eq:actm_demand_supply}
\end{equation}

\paragraph{The conservation law of a sink cell}
In a sink cell, there is no internal flow from the sink cell since it is the last cell on the road. The conservation law of vehicles in the sink cell is formulated in Equation~\ref{eq:con_sink},
\begin{equation}
    \begin{aligned}
        n_{r,s}^{-1}(t) &= n_{r,s}^{-1}(t-1) - \nu_{r,s}(t) - \phi_{r,s,*}(t) + f_{r,s}^{-2,-1}(t) \\
        \phi_{r,s,*}(t) &= \sum_{i \in \psi^{+}(s)}\phi_{r,s,i}(t) \\
        \phi_{r,s,i}(t) &= \hat{\phi} \left( \left\{ \mathcal{D}_{j,s}(t) \vert \forall j \in \psi^{-}(s) \right\}, \left\{ \mathcal{S}_{s,k}(t) \vert \forall k \in \psi^{+}(s) \right\} \right), \\
        & \qquad \qquad \qquad \forall r,s \in \mathbf{V}
    \end{aligned}
    \label{eq:con_sink}
\end{equation}
\noindent where $n_{r,s}^{-1}(t)$ is the vehicle number in the sink cell at time $t$, $f_{r,s}^{-2,-1}$ is the internal traffic flow from the second to last cell to the last cell (sink cell) on road $e_{r,s}$ at time $t$. The external flow of road $e_{r,s}$, $\phi_{r,s,*}$ will be distributed to downstream roads through the external flow regulator.

\paragraph{The conservation law of a general cell} In general cells, there is no external traffic flow, and trip source or sink. Hence we only consider the internal flows from on-ramps to mainlines, mainlines to mainlines and mainlines to off-ramps.

The conservation law of a general cell is modeled in Equation~\ref{eq:con_gen}, 
\begin{equation}
    \begin{aligned}
        n_{r,s}^{k}(t) &= n_{r,s}^{k}(t-1) + f_{r,s}^{k-1,k}(t) - f_{r,s}^{k,k+1}(t) + R_{r,s}^{k}(t) - S_{r,s}^{k}(t) \\
        f_{r,s}^{k,k+1}(t) &= \min \left\{ v_{r,s}^{max} \left(1- \beta_{r,s}^{k}(t)\right) \left(n_{r,s}^{k}(t) + \gamma R_{r,s}^k(t)\right), w_{r,s} \left(\hat{n}_{r,s} - n_{r,s}^{k+1}(t) - \gamma R_{r,s}^{k+1}(t) \right),  q_{r,s}^{max} \right\}, \\
        & \qquad \qquad \qquad \qquad \forall r,s \in \mathbf{V}, \ 2 \leq k \leq \lceil l_{r,s} / \delta_{r,s} \rceil - 1, \\
    \end{aligned}
    \label{eq:con_gen}
\end{equation}
\noindent where $w_{r,s}$ is the spillback speed of congestion at road $e_{r,s}$, $\beta_{r,s}^{k}(t)$ is the split ratio for off-ramp in cell $k$ on road $e_{r,s}$. $\gamma$ is the on-ramp flow blending parameter determined how much of the on-ramp flow is added to the mainstream before the mainstream arrives, which indicates the impact of mainstream from on-ramp flows, $\hat{n}_{r,s}$ is the jam density on road $e_{r,s}$.

It can be seen that the mainstream $f_{r,s}^{k,k+1}(t)$ should fulfill the following conditions.
\begin{align}
    f_{r,s}^{k,k+1}(t) + S_{r,s}^k(t) &\leq v_{r,s}^{max} \left(n_{r,s}^{k}(t) + \gamma R_{r,s}^k(t)\right) \label{eq:main_con1} \\
    f_{r,s}^{k,k+1}(t) &\leq w_{r,s} \left(\hat{n}_{r,s} - n_{r,s}^{k+1}(t) - \gamma R_{r,s}^{k+1}(t) \right) \label{eq:main_con2} \\
    f_{r,s}^{k,k+1}(t) &\leq q_{r,s}^{max}(t) \label{eq:main_con3}
\end{align}
Equation~\ref{eq:main_con1} restricts that the average speed of total vehicles left cell $k$ should under the free-flow speed. Equation~\ref{eq:main_con2} limits that the mainstream to the next cell should not exceed the supply of the next cell. Equation~\ref{eq:main_con3} ensures that the mainstream should not surpass the road capacity.

\paragraph{The conservation law of an off-ramp cell}
For an off-ramp cell, the conservation law is formulated in Equation~\ref{eq:con_down},
\begin{equation}
    \begin{aligned}
        n_{r'',s''}^{1}(t) &= n_{r'',s''}^{1}(t-1) - f_{r'',s''}^{1,2}(t) + S_{r,s}^{k}(t) \\
        S_{r,s}^{k}(t) &= \frac{\beta_{r,s}^{k}(t)}{1 - \beta_{r,s}^{k}(t)} f_{r,s}^{k,k+1}(t), \\
        & \forall r,s \in \mathbf{V}, \ k \leq \lceil l_{r,s} / \delta_{r,s} \rceil,
    \end{aligned}
    \label{eq:con_down}
\end{equation}
\noindent where $n_{r'',s''}^{1}(t)$ is the vehicle number in the first cell of road $e_{r'',s''}$, the off-ramp of road $e_{r,s}$. $S_{r,s}^{k}(t)$ is determined by the mainstream $f_{r,s}^{k,k+1}(t)$ and the split ratio of mainstream $\beta_{r,s}^{k}(t)$. In the meso-marco traffic modeling, the split ratio $\beta_{r,s}^{k}(t)$ varies according to the destination of vehicles in the cell.

\section{Demonstrators of agents}
\label{apx:demonstrator}
This section presents the detailed formulation for demonstrators on both freeways and local roads.
\paragraph{\textbf{The demonstrator of a ramp}}
ALINEA is a local ramp metering algorithm that aims to regulate the ramp flow on a local bottleneck. For an on-ramp, ALINEA takes the traffic occupancy as the input, and outputs the metered flow, which can be fulfilled by changing the green and red time of the signal light at a ramp. To better combine the ALINEA with the proposed simulator, we change the unit of decision variables and parameters in the original ALINEA method. The modified ALINEA controller is shown in Equation~\ref{eq:ALINEA}, 
\begin{equation}
    \rho_{r,s}^{k}(t+1) = \rho_{r,s}^{k}(t) + KI_{r,s}^{k} \left( \dot{n}_{r,s}^{k} - n_{r,s}^{k}(t) \right),
    \label{eq:ALINEA}
\end{equation}
\noindent where $\rho_{r,s}^{k}(t)$ is the metered rate of the $k$th ramp on the road $e_{r,s}$ defined in Equation~\ref{eq:con_up}. $n_{r,s}^{k}(t)$ is the vehicle number in the $k$th ramp on the road $e_{r,s}$. $\dot{n}_{r,s}^{k}$ is a threshold for the controlled vehicle number, which is usually the value of the critical density in the cell. $KI_{r,s}^{k}$ is a step-size parameter defined that how sensitive it reacts to traffic congestion. The larger the number is, the faster it delays vehicles on the ramp when the mainstream is congested, and vice versa. 

The ALINEA is an adaptive local control method that performs differently according to the level of congestion. This study quantizes the control rate of ALINEA using 
a step size of $\Delta u$.
By comparing to values in the threshold $\dot{n}_{r,s}^{k}$ and the vehicle number $n_{r,s}^{k}(t)$, the control action is calculated as follows:
\begin{equation}
    \mathbf{u}_t = \begin{cases}
    +\Delta \mathbf{u}, &\dot{n}_{r,s}^{k} - n_{r,s}^{k}(t) > 0 \\
    0, &\dot{n}_{r,s}^{k} - n_{r,s}^{k}(t) = 0 \\
    -\Delta \mathbf{u}, &\dot{n}_{r,s}^{k} - n_{r,s}^{k}(t) < 0 \\
    \end{cases}.
    \nonumber
\end{equation}

\paragraph{\textbf{The demonstrator of a perimeter}}
Gating is a local perimeter control method aiming to limit the traffic inflow into a congested urban area. For a perimeter, Gating takes the Total Time Spent (TTS) as the input, and outputs the inflow capacity at boundaries. Here the TTS refers to the number of vehicles in a region. The unit of decision variables and parameters in the original Gating method is also replaced to better fit this study. The modified Gating controller is shown in Equation~\ref{eq:Gating}:
\begin{equation}
    \Rho_{d}(t+1) = \Rho_{d}(t) - KP_{d}\left( N_{d}(t+1) - N_{d}(t) \right) + KI_{d} \left( \dot{N}_{d} - N_{d}(t) \right),
    \label{eq:Gating}
\end{equation}
\noindent where $\Rho_{d}(t)$ is the inflow rate of region $d$ in Equation~\ref{eq:bathtub_demand_supply}, $N_{d}(t)$ is the vehicle accumulation in region $d$, $\dot{N}_{d}(t)$ is a threshold for the controlled vehicle accumulation. Typically, the threshold is set as the critical vehicle accumulation of the MFD in the region. $KP_{d}$ and $KI_{d}$ are regulator parameters that weigh the impact on the inflow rate from the historical and critical vehicle accumulation.

Gating is also an adaptive local control method that responds differently according to the congestion level, and its control rate is a continuous variable in the system. To discretize the Gating method, the control action is defined as follows: 
\begin{equation}
    \mathbf{u}_t = \begin{cases}
    +\Delta \mathbf{u}, &KI_{d} \left( \dot{N}_{d} - N_{d}(t) \right) - KP_{d}\left( N_{d}(t+1) - N_{d}(t) \right) > 0 \\
    0, &KI_{d} \left( \dot{N}_{d} - N_{d}(t) \right) - KP_{d}\left( N_{d}(t+1) - N_{d}(t) \right) = 0 \\
    -\Delta \mathbf{u}, &KI_{d} \left( \dot{N}_{d} - N_{d}(t) \right) - KP_{d}\left( N_{d}(t+1) - N_{d}(t) \right) < 0 \\
    \end{cases}.
    \nonumber
\end{equation}
\end{document}